\documentclass[lettersize,journal]{IEEEtran}
\usepackage{amsmath,amsfonts}
\usepackage{algorithmic}
\usepackage{algorithm}
\usepackage{array}
\usepackage[caption=false,font=normalsize,labelfont=sf,textfont=sf]{subfig}
\usepackage{textcomp}
\usepackage{stfloats}
\usepackage{url}
\usepackage{verbatim}
\usepackage{graphicx}
\usepackage{xcolor}
\usepackage{cite}

\newcommand{\methodname}{CoopOcc}

\definecolor{svFree}{RGB}{255,255,255}
\definecolor{svBarrier}{RGB}{255,120,50}
\definecolor{svCar}{RGB}{0,150,245}
\definecolor{svPedestrian}{RGB}{255,0,0}
\definecolor{svTruck}{RGB}{160,32,240}
\definecolor{svSpeedLimitSign}{RGB}{255,255,0}
\definecolor{svVan}{RGB}{0,255,255}
\definecolor{svTrailer}{RGB}{135,60,0}
\definecolor{svTrafficCone}{RGB}{255,240,150}
\definecolor{svDrivable}{RGB}{255,0,255}
\definecolor{svTerrain}{RGB}{150,240,80}
\definecolor{svSidewalk}{RGB}{75,0,75}
\definecolor{svManmade}{RGB}{110,95,170}
\definecolor{svVegetation}{RGB}{0,175,0}
\newcommand{\classbox}[1]{\textcolor{#1}{\raisebox{0.05ex}{\rule{0.9ex}{0.9ex}}}\hspace{0.25ex}}
\newcommand{\freeclassbox}{\begingroup\setlength{\fboxsep}{0.02ex}\setlength{\fboxrule}{0.015ex}\raisebox{0.05ex}{\fcolorbox{black!15}{svFree}{\textcolor{svFree}{\rule{0.97ex}{0.97ex}}}}\endgroup\hspace{0.25ex}}

\begin{document}

\title{CoopOcc: Cooperative 3D Semantic Occupancy Perception via Multi-Agent and Cross-Modal Fusion}

\author{Zengle Zhu,
        Zhen Li,
        Tianyi Huai,
        Xinhu Zheng,~\IEEEmembership{Member,~IEEE,}
        Liuqing Yang,~\IEEEmembership{Fellow,~IEEE,}
        Rongqing Zhang,~\IEEEmembership{Member,~IEEE}%
\thanks{
Zengle Zhu, Zhen Li, Tianyi Huai, and Rongqing Zhang are with the Intelligent Transportation Thrust, The Hong Kong University of Science and Technology (Guangzhou), Guangzhou 510000, China (e-mail: \{zzhu622, zli619, thuai231\}@connect.hkust-gz.edu.cn; \mbox{rongqingz@hkust-gz.edu.cn}).

Xinhu Zheng is with the Internet of Things Thrust \& Intelligent Transportation Thrust, The Hong Kong University of Science and Technology (Guangzhou), Guangzhou 510000, China (e-mail: \mbox{xinhuzheng@hkust-gz.edu.cn}).

Liuqing Yang is with the Internet of Things Thrust \& Intelligent Transportation Thrust, The Hong Kong University of Science and Technology (Guangzhou), Guangzhou 510000, China, and also with the Department of Electronic and Computer Engineering, The Hong Kong University of Science and Technology, Hong Kong SAR, China (e-mail: \mbox{lqyang@ust.hk}).
}}

\markboth{Journal of \LaTeX\ Class Files,~Vol.~14, No.~8, August~2021}%
{Shell \MakeLowercase{\textit{et al.}}: A Sample Article Using IEEEtran.cls for IEEE Journals}


\maketitle

\begin{abstract}
Semantic occupancy perception has emerged as a promising paradigm for jointly modeling 3D scene geometry and semantic information. However, this unified representation cannot compensate for missing information caused by occlusion and sensor coverage. Multi-agent cooperation alleviates this limitation, but information from different agents is not equally useful across spatial regions. Meanwhile, feature representations from each agent form the basis of reliable multi-agent fusion. Cross-modal fusion constructs such representations by integrating complementary semantic and geometric information. Despite this complementarity, jointly establishing reliable 3D correspondence across agents and preserving modality-specific structures remains challenging. To address these limitations, we propose \methodname{}, which performs cooperative 3D semantic occupancy perception through multi-agent and cross-modal fusion. \methodname{} uses ego-query cross-agent attention (EQCA) to selectively aggregate aligned features across agents, and anchor-based bidirectional feature interaction (ABFI) for cross-modal exchange at matched coordinates. We further construct SemanticVVSim, a simulation dataset, and establish a unified benchmark. For simulation-to-real transfer, we propose confidence-aware hierarchical occupancy and relational distillation (CHORD). \methodname{} performs best overall among ten evaluated methods on SemanticVVSim. On Occ3D-nuScenes, CHORD improves performance using knowledge from SemanticVVSim. The code is available at \textcolor{black}{\url{https://github.com/LOTEAT/CoopOcc}}.
\end{abstract}


\section{Introduction}

\IEEEPARstart{P}{erception} is a cornerstone of autonomous driving, as it transforms sensor observations into structured scene understanding for safe and reliable decision-making~\cite{han2023collaborative}. Accurate perception requires autonomous vehicles to identify surrounding traffic participants and understand the 3D spatial layout of the surrounding environment, including drivable surfaces, sidewalks, terrain, and other scene regions. Object-centric 3D detection has long served as a primary formulation for autonomous driving perception, with a focus on the localization of traffic participants~\cite{zhou2018voxelnet,lang2019pointpillars,liu2023bevfusion}. It represents object instances using a sparse set of 3D bounding boxes and provides instance-level positions and spatial extents. However, this object-centric representation describes only part of the scene. Drivable surfaces, static structures, free space, and irregular non-object regions remain unmodeled~\cite{roldao2022survey,song2017sscnet}. Consequently, this representation provides limited spatial constraints for downstream planning and control. To address these representational limitations, 3D semantic occupancy perception represents the space around the ego vehicle using voxel-wise occupancy and semantic labels~\cite{cao2022monoscene,huang2023tpvformer,wei2023surroundocc}. This unified representation describes both object and non-object regions, making it a promising formulation for dense 3D scene understanding~\cite{tong2023sceneoccupancy,wang2023openoccupancy,tian2023occ3d}.

Dense semantic occupancy perception requires estimating occupancy throughout the entire 3D volume~\cite{roldao2022survey,tian2023occ3d}. Voxels outside the field of view or occluded by surrounding objects lack direct sensor observations~\cite{cao2022monoscene,li2023voxformer}. Such unobserved voxels are especially common at intersections or on congested roads. Occupancy estimates in these regions rely more heavily on learned scene priors. However, such priors may not provide reliable guidance for the current scene. Multi-agent cooperation alleviates this limitation by incorporating observations from spatially distributed agents~\cite{wang2020v2vnet,li2021disconet,xu2022v2xvit}. These agents provide complementary viewpoints~\cite{hu2022where2comm,song2024cohff,chen2026vogscp}. Through such cooperation, the occluded area may be directly observed by a nearby cooperating vehicle. Occupancy estimates for the corresponding voxels can rely more on direct sensor evidence and less on learned scene priors. Thus, broader observation coverage can reduce ambiguity in occupancy estimation and support more accurate dense semantic occupancy perception.

Effective use of these complementary viewpoints for dense semantic occupancy perception depends on how the local observations from each cooperating vehicle represent the surrounding scene. In autonomous driving, cameras capture rich appearance and semantic cues~\cite{cao2022monoscene,li2023voxformer,wei2023surroundocc}, while LiDAR provides direct 3D range measurements~\cite{behley2019semantickitti,zhu2021cylinder3d}. However, neither modality alone directly provides the dense, spatially explicit representation required for semantic occupancy perception. Cameras project the 3D scene onto 2D image planes, preserving dense visual information but leaving the 3D locations of observed content ambiguous~\cite{zhang2023occformer,huang2023tpvformer}. This ambiguity can affect the predicted occupancy states and semantic labels of multiple voxels. In contrast, LiDAR measurements provide reliable 3D support on observed scene surfaces, but sparsely sampling the space captures limited evidence for dense semantic perception~\cite{roldao2020lmscnet,yan2021js3cnet}. These sensing characteristics make the two modalities complementary: visual observations provide rich context, whereas LiDAR measurements provide sparse but explicit 3D anchors~\cite{pan2024coocc,zhang2024occfusion,duan2025sdgocc}.

Combining complementary camera and LiDAR observations from spatially distributed agents broadens observation coverage and supports dense 3D scene understanding, but multi-modal cooperative perception still faces three central challenges.
\subsubsection{Spatial Variation in Cross-Agent Complementarity}
For regions that are already sufficiently observed by the ego agent, features from other agents often overlap with existing information and provide limited additional benefit, potentially introducing redundancy~\cite{li2021disconet,hu2022where2comm}. In contrast, unobserved or insufficiently observed regions have a greater need for collaborative information, yet not every cooperating agent can provide effective complementary information due to differences in observation coverage and information quality. Consequently, both the demand for collaborative information and the useful contributions from different agents vary across spatial locations. For dense semantic occupancy perception over the surrounding 3D space, effectively assessing these spatially varying agent contributions is therefore crucial for collaborative fusion.

\subsubsection{Cross-Modal Representation Gap in Cross-Agent Fusion}
In cooperative perception, feature representations form the basis for subsequent cross-agent alignment and information aggregation~\cite{pan2024coocc,zhang2024occfusion}. Due to the substantial differences between cameras and LiDAR, each agent needs to first integrate local multi-modal observations before cross-agent cooperation. However, the two modalities represent 3D scenes in different ways. Camera features contain dense semantic information but lack explicit 3D positions~\cite{cao2022monoscene,huang2023tpvformer}. LiDAR features correspond to explicit 3D positions but remain sparse~\cite{behley2019semantickitti,zhu2021cylinder3d}. This difference makes reliable 3D correspondence across agents difficult to establish while preserving features from both modalities. Consequently, the construction of a unified scene representation and subsequent cross-agent fusion become more challenging.

\subsubsection{Lack of Comprehensive Dataset and Benchmark}
Comprehensive evaluation of multi-agent, multi-modal semantic occupancy perception requires diverse driving scenes and collaboration settings, including varied numbers, positions, and viewpoints of collaborating agents. However, existing semantic occupancy datasets mainly focus on single-agent perception~\cite{wang2023openoccupancy,tian2023occ3d}, while multi-agent datasets commonly target 3D object detection or BEV semantic segmentation~\cite{xu2022opv2v,li2022v2xsim,yu2022dairv2x,xu2023v2v4real,hao2024rcooper,zimmer2024tumtrafv2x}. This gap limits comprehensive evaluation and reliable comparison of multi-agent, multi-modal semantic occupancy perception across diverse scenes and collaboration conditions~\cite{song2024cohff,chen2026vogscp}.

To address these challenges, we propose \methodname{}, a cooperative 3D semantic occupancy perception framework that integrates observations through multi-agent and cross-modal fusion in a shared 3D space. \methodname{} uses ego-query cross-agent attention (EQCA) to selectively aggregate spatially corresponding features across agents. Within each agent, \methodname{} uses anchor-based bidirectional feature interaction (ABFI) to construct spatially consistent multi-modal features for cross-agent cooperation. ABFI exchanges camera and LiDAR information at coordinate-matched 3D locations. We further construct SemanticVVSim, a multi-agent, multi-modal simulation dataset, and establish a unified benchmark for semantic occupancy perception. We then investigate whether knowledge learned from SemanticVVSim can improve semantic occupancy perception in real scenes. However, differences in data distributions, sensor configurations, and semantic label spaces make direct transfer unreliable. Therefore, we introduce confidence-aware hierarchical occupancy and relational distillation (CHORD). The main contributions are summarized as follows.

{
\renewcommand{\labelenumi}{\alph{enumi})}
\begin{enumerate}
    \item To exploit complementary observations from spatially distributed vehicles while reducing interference from redundant or less relevant information, we propose \methodname{}, a multi-agent, multi-modal cooperative framework for semantic occupancy perception. Unlike methods that homogeneously aggregate features across agents, \methodname{} uses ego-query cross-agent attention (EQCA) to adaptively weight the contributions of different agents at each spatial location. EQCA aggregates only spatially corresponding features and dynamically adjusts the contribution of each agent at each location according to the ego representation, thereby selectively using complementary observations from other vehicles.

    \item To provide spatially consistent multi-modal representations for multi-agent cooperation, we introduce anchor-based bidirectional feature interaction (ABFI). Compared with multi-modal fusion in the single-agent setting, which primarily focuses on spatial correspondence across modalities, ABFI preserves explicit 3D physical relationships to support cross-view alignment and information aggregation across agents. Specifically, ABFI establishes bidirectional interaction between camera features and LiDAR voxel features to exchange complementary semantic and geometric information. Through this interaction, ABFI constructs a unified representation with explicit 3D geometric constraints and cross-modal spatial consistency.

    \item To support systematic study of multi-agent, multi-modal semantic occupancy perception, we construct SemanticVVSim, a multi-agent, multi-modal dataset for semantic occupancy perception. SemanticVVSim contains 61,625 synchronized multi-agent frames, comprising 255K LiDAR sweeps and 3.5M images, with dense 3D semantic occupancy annotations over 14 classes (e.g., free, car, pedestrian, drivable surface, sidewalk, manmade, and vegetation). The data cover 19 interaction scenarios across routine driving and safety-critical cases (e.g., hard braking, cut-ins, lane changes, and pedestrian jaywalking). The data also span five weather conditions (sunny, foggy, rainy, cloudy, and snowy) and 11 scene categories (e.g., city, town, university, highway, and mountain).

    \item To address differences in data distributions and semantic label spaces during simulation-to-real transfer, we introduce confidence-aware hierarchical occupancy and relational distillation (CHORD). CHORD selects occupancy regions for transfer according to teacher confidence. It distills latent features, spatial relations, binary occupancy outputs, and coarse semantic groups. CHORD avoids direct supervision between incompatible fine-grained semantic categories, enabling cross-domain transfer through a shared hierarchical occupancy representation.
\end{enumerate}
}

The rest of this paper is organized as follows: Section~\ref{sec:related_works} introduces the related works and explores previous research in related fields. Section~\ref{sec:methodology} details the architecture of \methodname{}. Section~\ref{sec:experiments} shows the performance results of \methodname{}. Section~\ref{sec:conclusion} concludes this paper.

\section{Related Works}
\label{sec:related_works}

\subsection{3D Semantic Occupancy Perception}

Perception is fundamental to autonomous driving because it enables vehicles to understand their surroundings and supports safe decision-making~\cite{han2023collaborative}. 3D object detection has long been a dominant task in autonomous driving perception~\cite{zhou2018voxelnet,lang2019pointpillars}. It identifies and localizes key traffic participants using 3D bounding boxes. However, this object-centric sparse representation provides limited descriptions of free space, road surfaces, and other non-object regions~\cite{roldao2022survey}. To obtain a dense scene-level 3D representation, 3D semantic occupancy perception models the occupancy state and semantic category of each voxel in the surrounding space~\cite{song2017sscnet,wang2023openoccupancy,tian2023occ3d}.

Early methods predict 3D semantic occupancy from depth maps or point clouds~\cite{song2017sscnet,guo2018vvnet,roldao2020lmscnet,li2020aicnet,yan2021js3cnet}. These inputs directly provide 3D locations, so the central problem is to recover complete voxel geometry and semantics from partial and sparse surface observations~\cite{roldao2022survey}. Research along this direction focuses on encoding sparse observations, modeling spatial context, and completing unobserved regions~\cite{song2017sscnet,roldao2020lmscnet,li2020aicnet}. However, the available sensor information limits this reconstruction process. Range measurements locate parts of visible scene surfaces but do not provide continuous, semantically rich observations throughout the voxel space~\cite{behley2019semantickitti,yan2021js3cnet}.

Therefore, camera-based methods have attracted increasing attention~\cite{cao2022monoscene,li2023voxformer}. Images provide dense appearance and semantic information across continuous 2D regions~\cite{wei2023surroundocc,zhang2023occformer}. This information enables models to use rich scene context to distinguish semantic categories~\cite{huang2023tpvformer,ma2024cotr}. Camera-based methods map multi-view image features into 3D representations~\cite{cao2022monoscene,li2023voxformer,wei2023surroundocc,huang2023tpvformer}. The representations bring dense visual information into semantic perception across the surrounding voxel space~\cite{tong2023sceneoccupancy,wang2023openoccupancy,tian2023occ3d}. Subsequent research further improves the organization of image features, strengthening scene-level reasoning while reducing the computational cost of the representation~\cite{huang2024gaussianformer,wang2024panoocc,huang2024selfocc}.

LiDAR-based methods provide spatially explicit but sparsely sampled 3D geometry, whereas camera-based methods provide dense appearance and semantic information~\cite{zhu2021cylinder3d,cao2022monoscene}. Therefore, research on 3D semantic occupancy perception has expanded from single-modal approaches to multi-modal camera-LiDAR methods~\cite{pan2024coocc,zhang2024occfusion}. They improve overall perception performance by combining complementary geometric and semantic information~\cite{liu2023bevfusion,duan2025sdgocc,ming2025occcylindrical}. These methods commonly align camera and LiDAR features in a shared 3D space to establish spatial correspondence for cross-modal fusion~\cite{pan2024coocc,zhang2024occfusion,liu2023bevfusion}. However, spatial alignment only identifies corresponding features and does not determine how semantic and geometric information should interact. Camera and LiDAR features still differ in spatial structure and observation density. Therefore, it remains challenging to maintain fine-grained spatial correspondence while enabling the two representations to update each other in multi-modal 3D semantic occupancy perception~\cite{duan2025sdgocc,ming2025occcylindrical}.

\subsection{Cooperative Perception}

Multi-modal fusion improves single-vehicle perception through complementary sensor cues but cannot overcome occlusion or limited sensing range~\cite{han2023collaborative}. Cooperative perception shares observations from spatially distributed agents, allowing complementary viewpoints to support scene understanding~\cite{chen2019fcooper,wang2020v2vnet}. Therefore, cooperative perception has become an important direction for extending sensing coverage and recovering information from occluded regions~\cite{li2021disconet,xu2022v2xvit}.

A primary design choice in cooperative perception concerns the level of information exchanged and fused across agents~\cite{han2023collaborative}. Raw-observation fusion shares images or point clouds and preserves relatively complete sensor information, but incurs high communication and processing costs~\cite{chen2019fcooper}. Output-level fusion exchanges only final outputs and reduces transmitted data, but discards useful intermediate information~\cite{liu2020when2com}. Intermediate feature fusion provides a compromise by exchanging learned scene representations~\cite{wang2020v2vnet,li2021disconet,xu2022v2xvit}. Consequently, research has increasingly shifted from choosing exchange levels to designing effective intermediate feature fusion mechanisms~\cite{hu2022where2comm,xu2023cobevt}.

Existing intermediate feature fusion methods use spatially adaptive fusion, transformer-based feature interaction, and communication-region selection to improve information exchange and feature aggregation across agents~\cite{wang2020v2vnet,li2021disconet,xu2022v2xvit,hu2022where2comm,xu2023cobevt,xiang2023hmvit}. However, most cooperative perception research focuses on 3D object detection and BEV semantic segmentation, which produce sparse object boxes and planar semantic maps, respectively~\cite{xu2022opv2v,li2022v2xsim,yu2022dairv2x,xu2023v2v4real,lu2023coalign,wei2023cobevflow,hao2024rcooper,zimmer2024tumtrafv2x}. In contrast, semantic occupancy perception estimates the occupancy state and semantic category of every voxel in the surrounding 3D space~\cite{song2024cohff,chen2026vogscp}. Therefore, this task places different requirements on cross-agent spatial correspondence and dense 3D feature aggregation. Large-scale benchmarks such as OpenOccupancy and Occ3D advance single-agent semantic occupancy perception, but cannot be used to study multi-vehicle cooperation~\cite{wang2023openoccupancy,tian2023occ3d}. Existing data resources remain insufficient for jointly studying multi-agent, multi-modal, dense 3D semantic occupancy perception or for systematic evaluation across diverse interaction scenarios and collaboration conditions.

\section{Methodology}
\label{sec:methodology}

To fully exploit complementary multi-modal observations captured by different agents from different viewpoints, we propose \methodname{} for cooperative 3D semantic occupancy perception. The remainder of this section describes the framework design and implementation in detail.

\subsection{Framework Overview}

\begin{figure*}[!t]
\centering
\includegraphics[width=\textwidth]{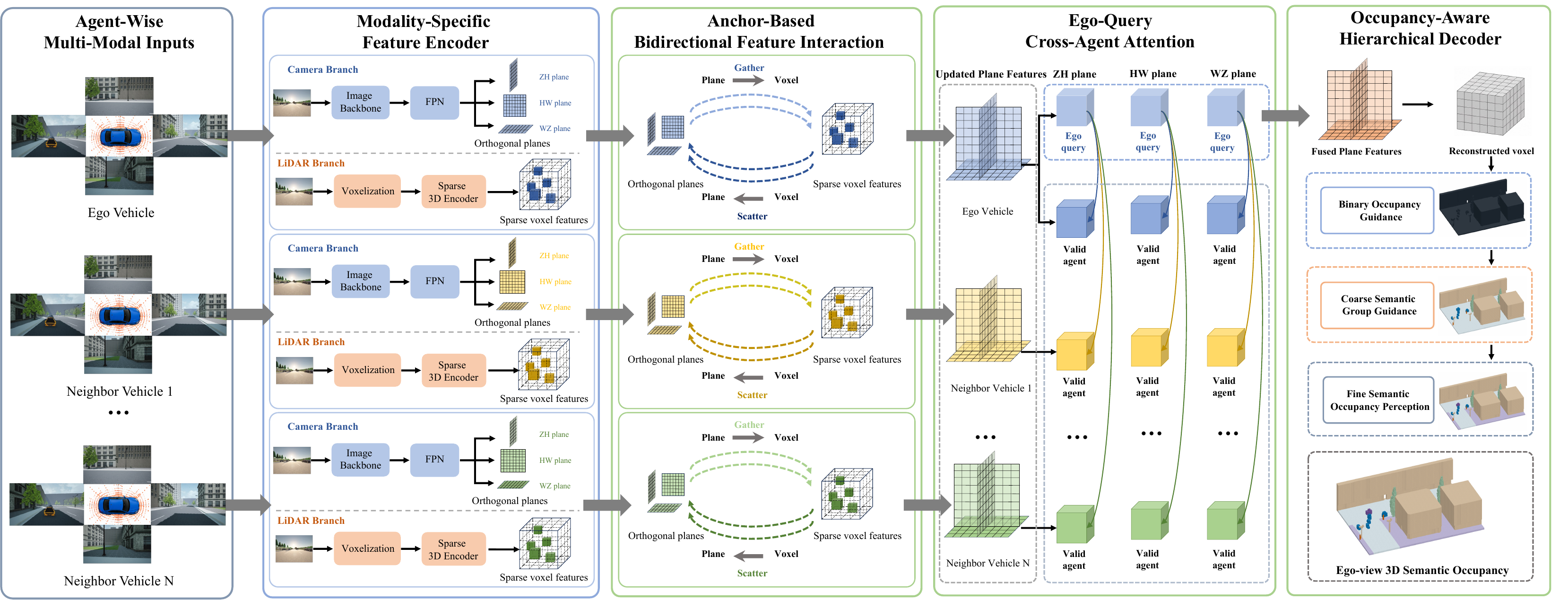}
\caption{Overview of the \methodname{} framework for cooperative 3D semantic occupancy perception. The framework combines modality-specific encoding, anchor-based bidirectional feature interaction, ego-query cross-agent attention, and hierarchical occupancy decoding.}
\label{fig:comaf_occ}
\end{figure*}

As illustrated in Figure~\ref{fig:comaf_occ}, \methodname{} is a unified framework for multi-agent, multi-modal 3D semantic occupancy perception. Given multi-view images and LiDAR point clouds collected by the ego agent and multiple cooperating agents, the framework constructs a multi-modal feature representation for each agent and establishes cross-agent spatial correspondence. The relevance of agent features to ego perception varies across spatial locations. Therefore, \methodname{} introduces ego-query cross-agent attention (EQCA). EQCA uses the ego feature as a query to assess spatially corresponding features from different agents. It adaptively adjusts the contribution of each agent at each location and selectively fuses complementary information into an ego-centered cooperative representation.

To support reliable cross-agent cooperation, modality-specific encoders first extract image and LiDAR features. Anchor-based bidirectional feature interaction (ABFI) then uses nonempty LiDAR voxels as shared 3D anchors to integrate camera semantics and LiDAR geometry at spatially corresponding locations. ABFI provides EQCA with spatially consistent multi-modal feature representations. Finally, the occupancy-aware hierarchical decoder converts the cooperative representation into a dense voxel grid and progressively predicts occupancy states and semantic classes.

\subsection{Modality-Specific Feature Encoder}
\label{sec:agent_encoding}

Camera images and LiDAR point clouds have distinct spatial structures. Therefore, we use a modality-specific feature encoder with separate camera and LiDAR branches. The coordinate frame of the ego LiDAR serves as the common spatial reference for all cooperative agents. The $x$, $y$, and $z$ axes denote the longitudinal, lateral, and vertical directions, respectively. Let $\mathcal{E}_{\mathrm{cam}}$ and $\mathcal{E}_{\mathrm{lidar}}$ denote the camera and LiDAR encoders, respectively. The same camera and LiDAR encoders are applied to all agents.

Let $\mathcal{A}=\{0,\ldots,A-1\}$ denote the set of valid agents, where agent $0$ is the ego vehicle. At each perception timestamp, agent $a$ provides multi-view images $\{\mathbf{I}_{a,v}\}_{v=1}^{V}$ and a LiDAR point cloud $\mathbf{P}_{a}$. The symbol $V$ denotes the number of camera views for each agent.

\subsubsection{Camera Encoder}
The camera encoder converts the multi-view images of agent $a$ into three orthogonal feature planes. A ResNet-101 backbone~\cite{he2016resnet} and a feature pyramid network~\cite{lin2017fpn} first extract a multi-scale feature pyramid $\mathbf{F}_{a,v}=\{\mathbf{F}^{(\ell)}_{a,v}\}_{\ell=1}^{L}$ from each camera view, where $\mathbf{F}^{(\ell)}_{a,v}\in\mathbb{R}^{h_\ell\times w_\ell\times d_\ell}$ denotes the feature map at pyramid level $\ell$. The symbols $h_\ell$, $w_\ell$, and $d_\ell$ denote the height, width, and channel dimension at this level, respectively. The encoder represents the 3D region around the ego vehicle in the common coordinate frame. This region is divided into $W$, $H$, and $Z$ positions along the $x$, $y$, and $z$ axes, respectively. Pairing the positions on two axes forms a 2D grid on each of the $xy$, $yz$, and $xz$ planes. The three grids contain $H\times W$, $Z\times H$, and $W\times Z$ locations, respectively. A learnable query vector represents each grid location. Let $\mathcal{P}=\{xy,yz,xz\}$ denote this set of planes.

A query location on a plane specifies two of the three spatial coordinates. The coordinate perpendicular to the plane remains unspecified. For example, a query at $(x,y)$ on the $xy$ plane specifies the horizontal position but not the height. The encoder samples several heights $z_k$ and obtains a set of 3D points $(x,y,z_k)$ along the vertical line through this query. Queries on the $yz$ and $xz$ planes use the same procedure along the $x$ and $y$ axes, respectively. The encoder uses the corresponding camera extrinsics and intrinsics to project these 3D points to 2D image coordinates for each camera. The projected points that fall within a camera view define valid reference locations. Multi-scale deformable attention~\cite{zhu2021deformabledetr} predicts sampling offsets and attention weights around these locations and samples from the multi-level feature pyramid for each valid camera. The sampled features are then aggregated across views to update the queries on the three orthogonal planes. The updated queries form the camera-derived orthogonal plane features
\begin{equation}
    \begin{aligned}
    \mathbf{C}_{a}
    &= \mathcal{E}_{\mathrm{cam}}\!\left(
    \{\mathbf{I}_{a,v}\}_{v=1}^{V}\right)\\
    &= \{\mathbf{C}^{xy}_{a},\mathbf{C}^{yz}_{a},\mathbf{C}^{xz}_{a}\}.
    \end{aligned}
    \label{eq:agent_camera_encoding}
\end{equation}
Let $d$ denote the feature dimension of each plane. Therefore, $\mathbf{C}^{xy}_{a}\in\mathbb{R}^{H\times W\times d}$, $\mathbf{C}^{yz}_{a}\in\mathbb{R}^{Z\times H\times d}$, and $\mathbf{C}^{xz}_{a}\in\mathbb{R}^{W\times Z\times d}$.

\subsubsection{LiDAR Encoder}
The LiDAR encoder preserves explicit 3D geometry through sparse voxel encoding. The point cloud from agent $a$ is first transformed into the ego-LiDAR coordinate frame using the sensor extrinsics and agent poses. The LiDAR encoder discretizes this region into a sparse voxel grid. The encoder retains only nonempty voxels, namely, voxels containing at least one point, for sparse feature extraction. We adopt a standard sparse voxel encoder following the LiDAR encoding architecture of BEVFusion~\cite{liu2023bevfusion}. The encoder applies mean pooling to the 3D point coordinates within each nonempty voxel and uses sparse 3D convolutions to aggregate features from neighboring voxels. Let $M_a$ denote the number of nonempty voxels in the sparse output for agent $a$. Each nonempty voxel $m$ has a discrete 3D grid index $\boldsymbol{\xi}_{a,m}=(\xi^{x}_{a,m},\xi^{y}_{a,m},\xi^{z}_{a,m})$ and a feature vector $\mathbf{v}_{a,m}$. The three index components identify the voxel location along the $x$, $y$, and $z$ axes of the sparse feature grid, respectively. We represent the resulting sparse voxel features as
\begin{equation}
    \begin{aligned}
    \mathbf{V}_{a}
    &= \mathcal{E}_{\mathrm{lidar}}\!\left(
    \mathbf{P}_{a}\right)\\
    &= \{(\boldsymbol{\xi}_{a,m},\mathbf{v}_{a,m})\}_{m=1}^{M_a}.
    \end{aligned}
    \label{eq:agent_lidar_encoding}
\end{equation}

The modality-specific feature encoder produces camera-plane features $\mathbf{C}_{a}$ and sparse LiDAR voxel features $\mathbf{V}_{a}$ for each agent. The camera branch organizes visual features on three dense orthogonal planes, whereas the LiDAR branch represents geometric features at nonempty 3D voxels.
\subsection{Anchor-Based Bidirectional Feature Interaction}
Camera-plane features and LiDAR voxel features describe the same 3D region through complementary representations. Camera-plane features encode dense visual information on three orthogonal planes. However, each plane aggregates 3D information along one spatial axis. This aggregation loses precise positional distinctions along that axis and prevents an explicit feature representation for every 3D voxel. In contrast, the LiDAR encoder preserves features at explicit 3D voxel locations, but only at spatially sparse sites supported by LiDAR measurements. Directly converting either representation into the other requires dense 3D lifting or discard their complementary structural properties.

\begin{figure}[!t]
\centering
\includegraphics[width=\columnwidth]{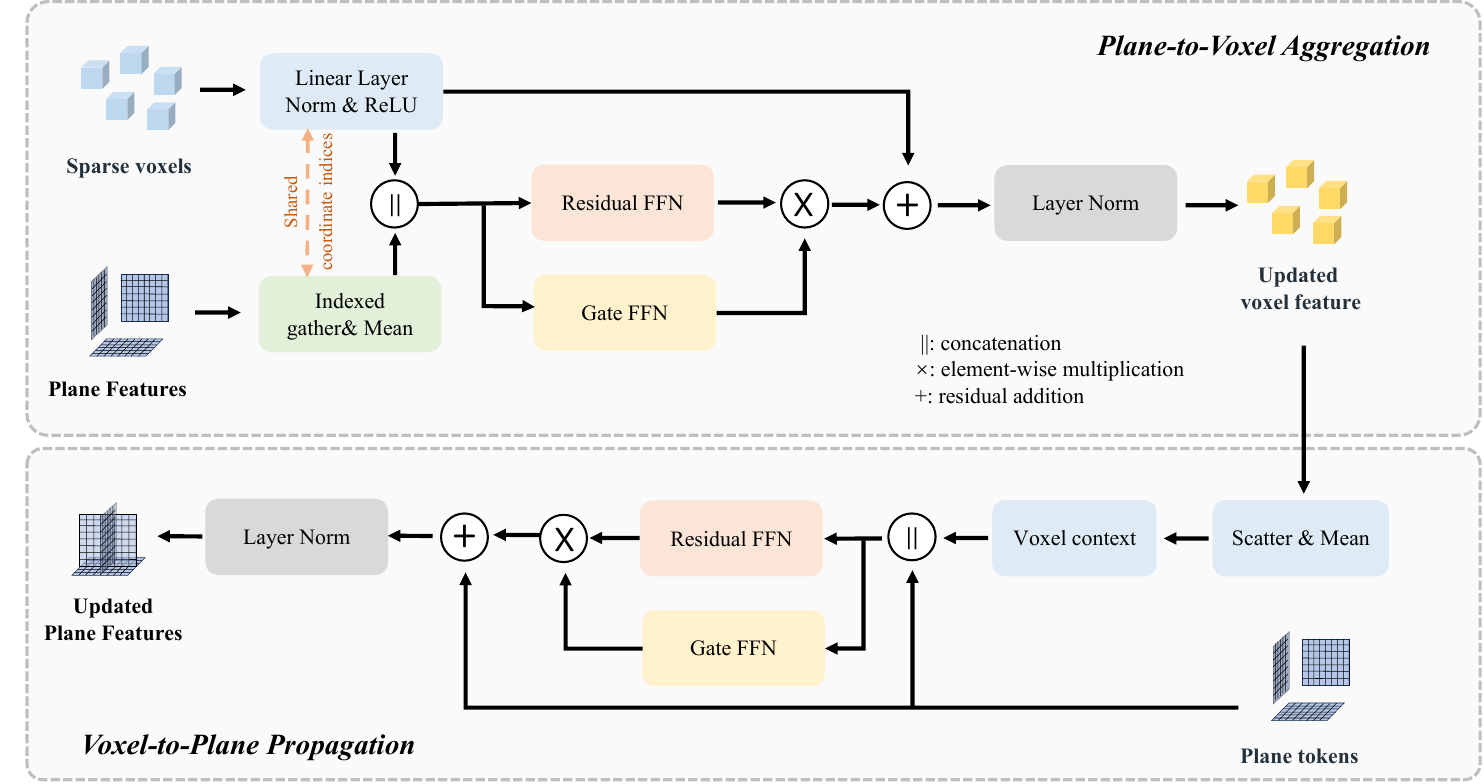}
\caption{Overview of anchor-based bidirectional feature interaction (ABFI).}
\label{fig:abfi}
\end{figure}

To bridge this representational gap, we propose anchor-based bidirectional feature interaction (ABFI), as shown in Figure~\ref{fig:abfi}. Rather than converting either modality into a common dense volume, ABFI uses sparse LiDAR-supported 3D anchors as an interface between the two representations. This interface injects plane context into explicit 3D locations and returns geometry-conditioned updates to the compact plane representation through coordinate-aligned bidirectional pathways.

\subsubsection{Plane-to-Voxel Aggregation}

ABFI uses nonempty LiDAR voxels as shared 3D anchors. Plane-to-voxel aggregation is the first of two complementary interaction directions. It gathers features from coordinate-matched locations on the three camera planes for each anchor and applies a gated residual update to the voxel feature. Although the sparse voxel grid and the plane grids share the same resolution along the $x$ and $y$ axes, the resolution along the $z$ axis differs. Therefore, the $xy$-plane index follows the $x$ and $y$ voxel coordinates directly, whereas the $yz$- and $xz$-plane indices rescale the $z$ coordinate before flattening. Let $\rho_z(\cdot)$ map a $z$-axis index from the sparse voxel grid to the coordinate-aligned $z$-axis index on the plane grids. We flatten $\mathbf{C}^{xy}_{a}$, $\mathbf{C}^{yz}_{a}$, and $\mathbf{C}^{xz}_{a}$ into sequences of $HW$, $ZH$, and $WZ$ tokens, respectively. Each token has $d$ feature channels. Using the voxel-grid index $\boldsymbol{\xi}_{a,m}=(\xi^{x}_{a,m},\xi^{y}_{a,m},\xi^{z}_{a,m})$, we define the flattened token indices as

\begin{equation}
    \begin{aligned}
    i^{xy}_{a,m}
    &= \xi^{y}_{a,m}W+\xi^{x}_{a,m},\\
    i^{yz}_{a,m}
    &= \rho_z(\xi^{z}_{a,m})H+\xi^{y}_{a,m},\\
    i^{xz}_{a,m}
    &= \xi^{x}_{a,m}Z+\rho_z(\xi^{z}_{a,m}).
    \end{aligned}
    \label{eq:voxel_plane_indices}
\end{equation}
The scalar indices $i^{xy}_{a,m}$, $i^{yz}_{a,m}$, and $i^{xz}_{a,m}$ select coordinate-matched tokens from the $xy$, $yz$, and $xz$ planes, respectively. These indices form a sparse correspondence graph between each LiDAR anchor and the three camera planes.

Plane-to-voxel aggregation uses geometry-based routing to gather the three coordinate-matched camera tokens for each voxel. For voxel $m$, Eq.~\eqref{eq:voxel_plane_indices} determines one coordinate-matched token index on each camera-derived plane. The plane-to-voxel routing operator $\mathcal{R}_{P\rightarrow V}$ gathers the three indexed tokens and averages them to form the camera context $\mathbf{c}_{a,m}$:
\begin{equation}
    \mathbf{c}_{a,m}
    =\mathcal{R}_{P\rightarrow V}(\mathbf{C}_{a})_m
    =\frac{1}{3}\sum_{p\in\mathcal{P}}
    \mathbf{C}^{p}_{a}[i^{p}_{a,m}].
    \label{eq:plane_to_voxel_routing}
\end{equation}
Before the gated update, a linear projection followed by layer normalization and ReLU maps each LiDAR feature $\mathbf{v}_{a,m}$ to the $d$-channel feature $\mathbf{u}_{a,m}$, which matches the channel dimension of the camera context $\mathbf{c}_{a,m}$.
Coordinate routing first delivers a camera context $\mathbf{c}_{a,m}$ to each nonempty LiDAR anchor. ABFI retains the aligned LiDAR feature $\mathbf{u}_{a,m}$ as the anchor state, and the routed camera context conditions a residual correction. This residual design injects camera context into sparse voxel features while preserving explicit 3D support. A channel-wise gate further controls the strength of this correction for each feature channel. Two separate feed-forward networks (FFNs) generate the residual correction and the channel-wise gate:
\begin{equation}
    \begin{aligned}
    \boldsymbol{\delta}^{v}_{a,m}
    &=\operatorname{FFN}^{v}_{\delta}
    ([\mathbf{u}_{a,m}\Vert\mathbf{c}_{a,m}]),\\
    \mathbf{g}^{v}_{a,m}
    &=\sigma\!\left(
    \operatorname{FFN}^{v}_{g}
    ([\mathbf{u}_{a,m}\Vert\mathbf{c}_{a,m}])\right),\\
    \widetilde{\mathbf{v}}_{a,m}
    &=\operatorname{LN}_{v}\!\left(
    \mathbf{u}_{a,m}
    +\mathbf{g}^{v}_{a,m}
    \odot\boldsymbol{\delta}^{v}_{a,m}\right).
    \end{aligned}
    \label{eq:plane_to_voxel_aggregation}
\end{equation}
The residual FFN $\operatorname{FFN}^{v}_{\delta}$ generates a candidate feature correction. The gate FFN $\operatorname{FFN}^{v}_{g}$ produces channel-wise update weights after the sigmoid function. The two FFNs use separate parameters. Thus, the residual branch determines the update content, and the gate branch controls the update strength. The resulting feature $\widetilde{\mathbf{v}}_{a,m}$ represents a camera-conditioned LiDAR anchor. The symbols $\sigma$, $\Vert$, and $\odot$ denote the sigmoid function, concatenation, and element-wise multiplication, respectively.

\subsubsection{Voxel-to-Plane Propagation}

Voxel-to-plane propagation maps the updated anchor features back to the corresponding plane tokens through normalized scatter, followed by a second gated residual update. Both directions use the same sparse set of coordinate-derived anchor-to-token links. For each anchor, the ego-LiDAR coordinates determine one spatially corresponding token on every plane.

Voxel-to-plane propagation begins with reverse routing over the same coordinate correspondences. The routing operator propagates the updated anchor features to the three camera-derived planes. Several voxels can map to one plane token because each plane omits one spatial axis. Let $\widetilde{\mathbf{V}}_{a}=\{\widetilde{\mathbf{v}}_{a,m}\}_{m=1}^{M_a}$ denote the updated LiDAR anchor features from agent $a$. The indicator function $\mathbb{1}[\cdot]$ equals one when the enclosed condition holds and zero otherwise. For token $j$ on plane $p\in\mathcal{P}$, $n^{p}_{a,j}$ denotes the number of updated anchors routed to this token. The feature $\mathbf{s}^{p}_{a,j}$ denotes the mean feature of these anchors, and $r^{p}_{a,j}$ indicates whether the token receives support from at least one anchor. We denote this normalized scatter operation by $\mathcal{R}_{V\rightarrow P}^{p}$:
\begin{equation}
    \begin{aligned}
    n^{p}_{a,j}
    &=\sum_{m=1}^{M_a}
    \mathbb{1}\!\left[i^{p}_{a,m}=j\right],\\
    \mathbf{s}^{p}_{a,j}
    &=\mathcal{R}_{V\rightarrow P}^{p}
    (\widetilde{\mathbf{V}}_{a})_j\\
    &=\frac{1}{\max(1,n^{p}_{a,j})}
    \sum_{m=1}^{M_a}
    \mathbb{1}\!\left[i^{p}_{a,m}=j\right]
    \widetilde{\mathbf{v}}_{a,m},\\
    r^{p}_{a,j}
    &=\mathbb{1}\!\left[n^{p}_{a,j}>0\right].
    \end{aligned}
    \label{eq:voxel_to_plane_routing}
\end{equation}
The routing operator $\mathcal{R}_{V\rightarrow P}^{p}$ averages all updated anchor features mapped to the same plane token. The support indicator suppresses the cross-modal residual when no LiDAR voxel maps to that token. Voxel-to-plane propagation treats $\mathbf{C}^{p}_{a}[j]$ as the receiving plane feature and $\mathbf{s}^{p}_{a,j}$ as the routed LiDAR context. The propagation stage estimates a separate residual feature and channel-wise gate:
\begin{equation}
    \begin{aligned}
    \boldsymbol{\delta}^{p}_{a,j}
    &=\operatorname{FFN}^{p}_{\delta}
    ([\mathbf{C}^{p}_{a}[j]\Vert\mathbf{s}^{p}_{a,j}]),\\
    \mathbf{g}^{p}_{a,j}
    &=\sigma\!\left(
    \operatorname{FFN}^{p}_{g}
    ([\mathbf{C}^{p}_{a}[j]\Vert\mathbf{s}^{p}_{a,j}])\right),\\
    \widetilde{\mathbf{C}}^{p}_{a}[j]
    &=\operatorname{LN}_{p}\!\left(
    \mathbf{C}^{p}_{a}[j]
    +r^{p}_{a,j}\,\mathbf{g}^{p}_{a,j}
    \odot\boldsymbol{\delta}^{p}_{a,j}\right).
    \end{aligned}
    \label{eq:voxel_to_plane_propagation}
\end{equation}
For each plane $p$, the residual FFN $\operatorname{FFN}^{p}_{\delta}$ generates a candidate feature correction. The gate FFN $\operatorname{FFN}^{p}_{g}$ produces channel-wise update weights after the sigmoid function. The two FFNs use separate parameters. The $xy$, $yz$, and $xz$ planes use separate residual and gate FFNs. The binary indicator $r^{p}_{a,j}$ equals one when at least one updated LiDAR anchor maps to token $j$, and zero otherwise. It suppresses the residual update for plane tokens without a mapped LiDAR anchor.

The aggregation and propagation stages use separate transformations but follow the same gated residual structure. Together, the two routing directions form a closed-loop plane-to-voxel-to-plane interaction path over the same sparse correspondence graph. Plane-to-voxel aggregation uses camera context to update each explicit 3D anchor. Voxel-to-plane propagation returns the updated anchor features to the compact plane representation used by cross-agent fusion and occupancy decoding. Coordinate routing determines where feature exchange occurs. Learned transformations and gates determine the update content and strength.

\subsection{Ego-Query Cross-Agent Attention}

For each agent $a$, ABFI produces the updated orthogonal-plane representation $\widetilde{\mathbf{C}}_{a}$. After flattening the spatial dimensions, let $\widetilde{\mathbf{C}}^{p}_{a}\in\mathbb{R}^{N_p\times d}$ denote the token sequence on plane $p$, where $N_{xy}=HW$, $N_{yz}=ZH$, and $N_{xz}=WZ$. At spatial index $j$, the corresponding tokens from all $A$ agents form the agent stack $\widetilde{\mathbf{C}}^{p}_{\mathcal{A}}[j]\in\mathbb{R}^{A\times d}$. The camera and LiDAR encoders construct all agent representations on the same grids in the ego-LiDAR coordinate frame. Therefore, tokens $\widetilde{\mathbf{C}}^{p}_{a}[j]$ with the same plane $p$ and spatial index $j$ describe the same spatial column across agents. This shared indexing constructs explicit cross-agent correspondence and avoids unrestricted matching between unrelated spatial locations. However, geometric correspondence alone does not determine the contribution of each agent. Occlusion, sensing range, and viewpoint cause spatially corresponding features to contain different information. A fixed aggregation rule cannot adapt the contribution of each agent to the current location and ego context.

\begin{figure}[!t]
\centering
\includegraphics[width=\columnwidth]{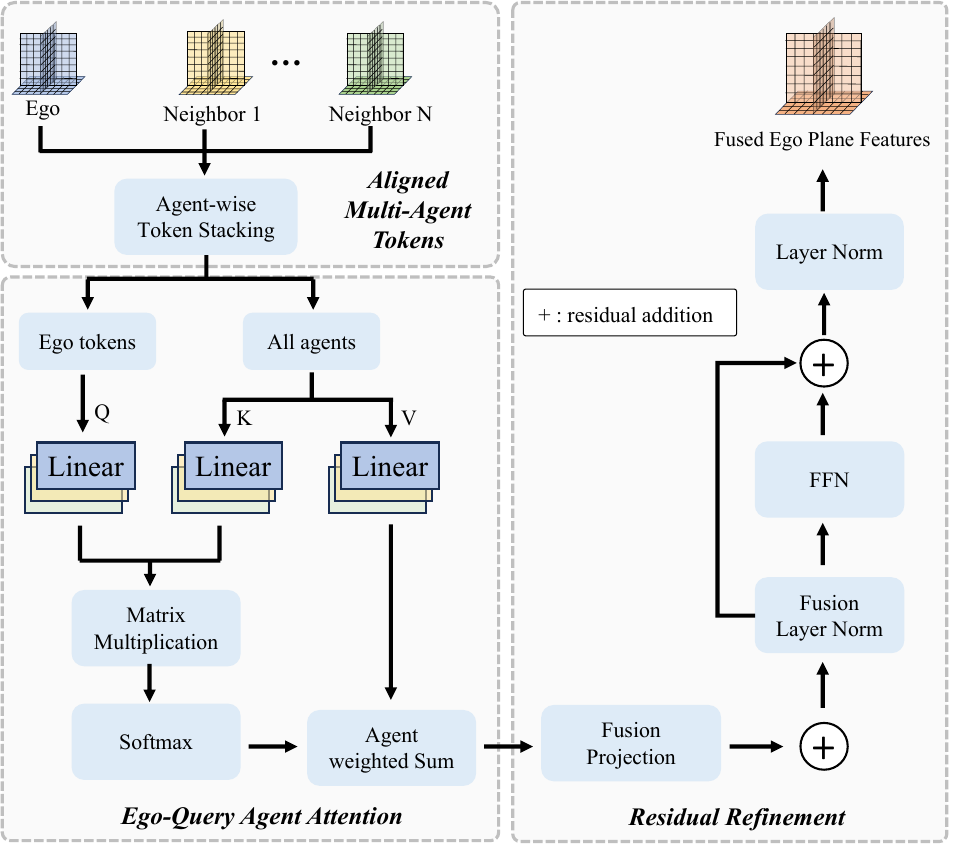}
\caption{Overview of ego-query cross-agent attention (EQCA).}
\label{fig:eqca}
\end{figure}

To address this limitation, we propose ego-query cross-agent attention (EQCA) through three coupled design choices, as illustrated in Figure~\ref{fig:eqca}. First, the common ego-coordinate indexing restricts the candidate set at $(p,j)$ to spatially corresponding tokens $\{\widetilde{\mathbf{C}}^{p}_{a}[j]\}_{a\in\mathcal{A}}$. Second, the ego token serves as the sole query, which makes the fusion asymmetric and conditions agent aggregation on the ego representation. Third, attention operates only along the agent dimension and is applied independently to the three compact orthogonal planes. These choices separate geometric correspondence construction from content-dependent agent aggregation while preserving an ego-centered plane representation.

For plane $p$ and spatial index $j$, EQCA first projects the updated ego token into a query and the same-index token from each agent into a key and value:
\begin{equation}
    \begin{aligned}
    \mathbf{q}^{p}_{0,j}
    &=\mathbf{W}^{p}_{q}\widetilde{\mathbf{C}}^{p}_{0}[j],\\
    \mathbf{k}^{p}_{a,j}
    &=\mathbf{W}^{p}_{k}\widetilde{\mathbf{C}}^{p}_{a}[j],\\
    \mathbf{v}^{p}_{a,j}
    &=\mathbf{W}^{p}_{v}\widetilde{\mathbf{C}}^{p}_{a}[j].
    \end{aligned}
    \label{eq:eqca_projection}
\end{equation}
The projected tokens are aggregated through attention along the agent dimension:
\begin{equation}
    \begin{aligned}
    \alpha^{p}_{a,j}
    &=\operatorname{softmax}_{a\in\mathcal{A}}\!\left(
    \frac{(\mathbf{q}^{p}_{0,j})^{\top}\mathbf{k}^{p}_{a,j}}
    {\sqrt{d}}\right),\\
    \mathbf{o}^{p}_{j}
    &=\sum_{a\in\mathcal{A}}\alpha^{p}_{a,j}
    \mathbf{v}^{p}_{a,j}.
    \end{aligned}
    \label{eq:eqca_attention}
\end{equation}
Here, $d$ is the feature dimension. At each location $(p,j)$, the ego token serves as the query, while the spatially corresponding tokens from all agents serve as keys and values. The softmax normalizes the attention weights across agents. EQCA therefore forms the ego update from a weighted combination of aligned agent features without mixing tokens from different spatial locations.

The attention output updates the ego token through an output projection and two residual refinement stages:
\begin{equation}
    \begin{aligned}
    \widehat{\mathbf{C}}^{p}_{0}[j]
    &=\operatorname{LN}^{p}_{1}\!\left(
    \widetilde{\mathbf{C}}^{p}_{0}[j]
    +\mathbf{W}^{p}_{o}\mathbf{o}^{p}_{j}\right),\\
    \overline{\mathbf{C}}^{p}_{0}[j]
    &=\operatorname{LN}^{p}_{2}\!\left(
    \widehat{\mathbf{C}}^{p}_{0}[j]
    +\operatorname{FFN}^{p}\!\left(
    \widehat{\mathbf{C}}^{p}_{0}[j]\right)\right).
    \end{aligned}
    \label{eq:eqca_update}
\end{equation}
The first residual stage injects the aggregated cross-agent feature into the ego token. The second stage refines the fused feature through a feed-forward network. EQCA derives the query only from the ego token and constructs the key-value set from all valid agents, including the ego vehicle. This formulation balances the ego observation and collaborator observations at each spatial location. The resulting ego-centered representation $\overline{\mathbf{C}}_{0}=\{\overline{\mathbf{C}}^{p}_{0}\}_{p\in\mathcal{P}}$ is passed directly to the occupancy decoder.

\subsection{Occupancy-Aware Hierarchical Decoder}

Dense semantic occupancy perception requires an occupancy state and a semantic class for every voxel in a dense 3D grid. Simply reconstructing voxel features and mapping each voxel directly to free space or one of the fine-grained semantic classes treats all output labels at a single level. This flat formulation does not explicitly use the shared occupied state of the fine-grained semantic classes or the group-level relations among semantically related classes. Therefore, the decoder reconstructs voxel-level features from the fused planes and organizes occupancy estimation as a progressive refinement process from binary occupancy to coarse semantic groups and fine-grained classes.

\begin{figure}[!t]
\centering
\includegraphics[width=\columnwidth]{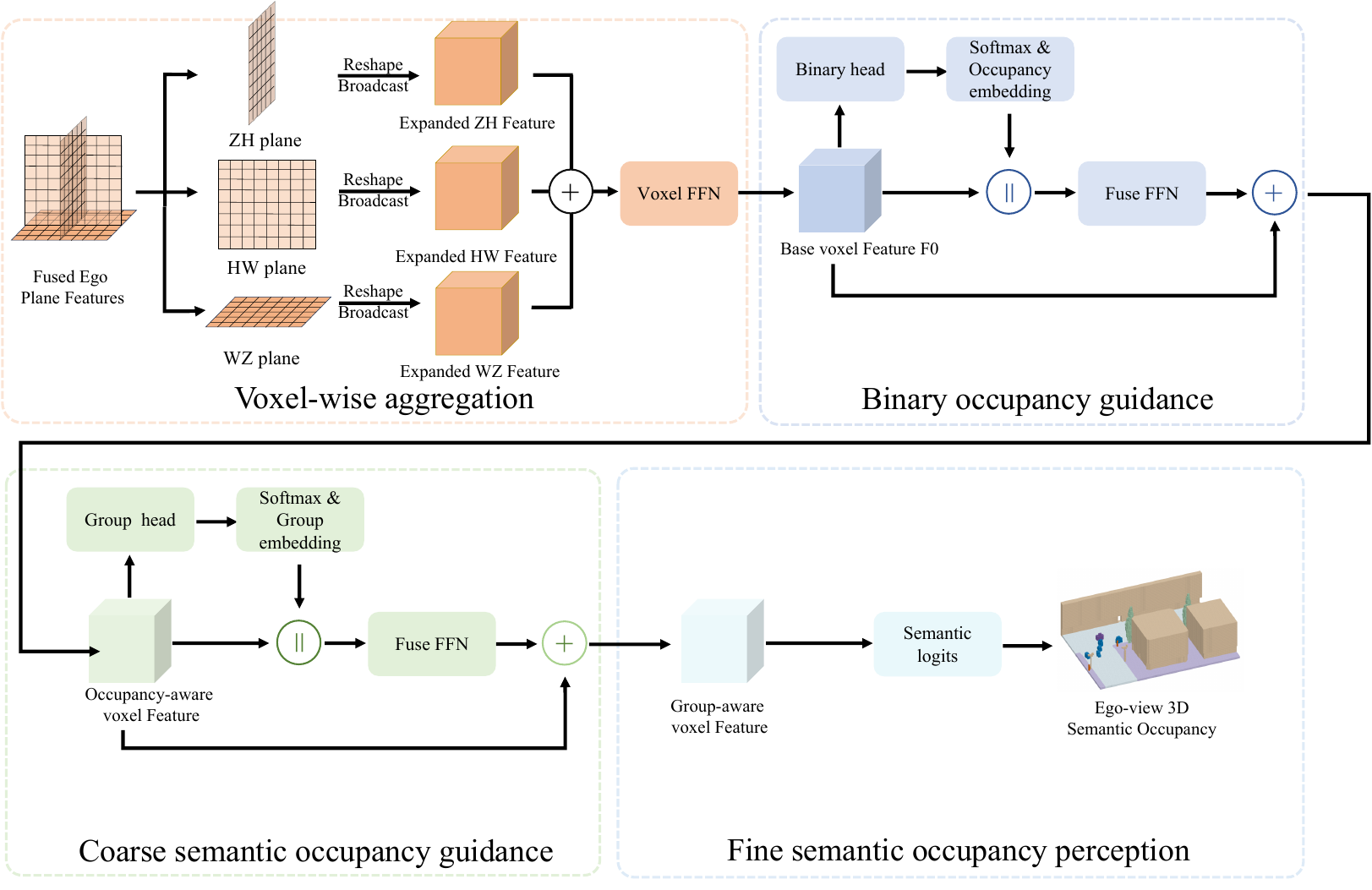}
\caption{Overview of the occupancy-aware hierarchical decoder.}
\label{fig:decoder}
\end{figure}

The decoder follows a coarse-to-fine classification hierarchy, as depicted in Figure~\ref{fig:decoder}. It first estimates binary occupancy to capture the basic geometric structure and then predicts six coarse semantic groups to provide group-level context. Based on these intermediate outputs, the final level distinguishes fine-grained semantic classes. Instead of propagating hard labels across levels, the decoder embeds each preceding soft distribution into the corresponding voxel features. This differentiable design preserves classification uncertainty and provides structured context for subsequent fine-grained classification.

Let $\mathcal{G}$ denote the set of all $WHZ$ voxels in the dense output grid. Each voxel $\nu\in\mathcal{G}$ corresponds to one feature location on each orthogonal plane. Let $(\nu_x,\nu_y,\nu_z)$ denote the grid location of voxel $\nu$. The pairs $(\nu_x,\nu_y)$, $(\nu_y,\nu_z)$, and $(\nu_x,\nu_z)$ directly index the corresponding features on the $xy$, $yz$, and $xz$ planes, respectively. The three indexed features are summed and passed through the decoder FFN to obtain the voxel feature:
\begin{equation}
    \mathbf{f}^{0}_{\nu}
    =\operatorname{FFN}_{\mathrm{dec}}\!\left(
    \sum_{p\in\mathcal{P}}
    \overline{\mathbf{C}}^{p}_{0}[\iota_p(\nu)]\right).
    \label{eq:plane_voxel_reconstruction}
\end{equation}
Here, $\iota_p(\nu)$ returns the flattened token index corresponding to voxel $\nu$ on plane $p$, and $\mathbf{f}^{0}_{\nu}$ is the reconstructed voxel feature.

The first level predicts free and occupied logits. Their soft probabilities form a weighted combination of two learned state embeddings. This binary context updates the voxel feature before group classification:
\begin{equation}
    \begin{aligned}
    \boldsymbol{\ell}^{\mathrm{bin}}_{\nu}
    &=h_{\mathrm{bin}}(\mathbf{f}^{0}_{\nu}),\\
    \boldsymbol{\pi}^{\mathrm{bin}}_{\nu}
    &=\operatorname{softmax}(\boldsymbol{\ell}^{\mathrm{bin}}_{\nu}),\\
    \mathbf{e}^{\mathrm{bin}}_{\nu}
    &=(\boldsymbol{\pi}^{\mathrm{bin}}_{\nu})^{\top}
    \mathbf{E}_{\mathrm{bin}},\\
    \mathbf{f}^{\mathrm{bin}}_{\nu}
    &=\mathbf{f}^{0}_{\nu}+\phi_{\mathrm{bin}}
    ([\mathbf{f}^{0}_{\nu}\Vert\mathbf{e}^{\mathrm{bin}}_{\nu}]).
    \end{aligned}
    \label{eq:binary_conditioning}
\end{equation}
Here, $h_{\mathrm{bin}}$ denotes the learned binary classification head, $\mathbf{E}_{\mathrm{bin}}\in\mathbb{R}^{2\times d}$ contains the learned free and occupied state embeddings, and $\phi_{\mathrm{bin}}$ is a two-layer feed-forward fusion network with a Softplus activation. We use a residual update to preserve the reconstructed voxel feature and introduce binary occupancy context as a learned correction for subsequent group classification. Using the soft distribution keeps this conditioning differentiable and retains uncertainty between the two occupancy states.

The second level predicts a coarse semantic group from the binary-conditioned feature. We organize the occupied classes into six groups that describe vehicles, vulnerable road users, traffic objects, flat surfaces, structures, and nature. The group distribution is embedded and returned to the voxel feature before fine classification:
\begin{equation}
    \begin{aligned}
    \boldsymbol{\ell}^{\mathrm{grp}}_{\nu}
    &=h_{\mathrm{grp}}(\mathbf{f}^{\mathrm{bin}}_{\nu}),\\
    \boldsymbol{\pi}^{\mathrm{grp}}_{\nu}
    &=\operatorname{softmax}(\boldsymbol{\ell}^{\mathrm{grp}}_{\nu}),\\
    \mathbf{e}^{\mathrm{grp}}_{\nu}
    &=(\boldsymbol{\pi}^{\mathrm{grp}}_{\nu})^{\top}
    \mathbf{E}_{\mathrm{grp}},\\
    \mathbf{f}^{\mathrm{grp}}_{\nu}
    &=\mathbf{f}^{\mathrm{bin}}_{\nu}+\phi_{\mathrm{grp}}
    ([\mathbf{f}^{\mathrm{bin}}_{\nu}\Vert\mathbf{e}^{\mathrm{grp}}_{\nu}]),\\
    \boldsymbol{\ell}^{\mathrm{sem}}_{\nu}
    &=h_{\mathrm{sem}}(\mathbf{f}^{\mathrm{grp}}_{\nu}).
    \end{aligned}
    \label{eq:group_semantic_conditioning}
\end{equation}
Here, $h_{\mathrm{grp}}$ and $h_{\mathrm{sem}}$ denote the learned classification heads for coarse semantic groups and fine-grained classes, respectively. The matrix $\mathbf{E}_{\mathrm{grp}}\in\mathbb{R}^{6\times d}$ contains the learned group embeddings, and $\phi_{\mathrm{grp}}$ is a two-layer feed-forward fusion network with a Softplus activation. We apply the same residual update at the group level to preserve the binary-conditioned feature and introduce coarse semantic context as a learned correction for fine classification. The three logits are supervised with binary occupancy, coarse-group, and fine semantic targets derived from the voxel labels. This coarse-to-fine hierarchy provides progressively refined context from occupancy states to semantic groups and fine-grained classes. The preceding outputs guide each finer level with structured context while preserving the underlying voxel representation.

We also supervise each fused orthogonal plane before voxel reconstruction. For a plane token, its binary target is occupied if any valid voxel along the omitted axis is occupied. We use binary rather than semantic plane supervision because each plane token represents a voxel column that can contain multiple semantic classes. Binary occupancy provides an unambiguous target by indicating whether any valid voxel in the column is occupied. Separate two-class heads predict these projected targets on the $xy$, $yz$, and $xz$ planes. The plane losses combine cross-entropy and Lovasz terms. The complete supervised objective is
\begin{equation}
    \mathcal{L}_{\mathrm{sup}}
    =\mathcal{L}_{\mathrm{sem}}+\mathcal{L}_{\mathrm{grp}}
    +\mathcal{L}_{\mathrm{bin}}
    +\sum_{p\in\mathcal{P}}\mathcal{L}^{p}_{\mathrm{plane}}.
    \label{eq:hierarchical_supervision}
\end{equation}
Here, $\mathcal{L}_{\mathrm{sem}}$, $\mathcal{L}_{\mathrm{grp}}$, and $\mathcal{L}_{\mathrm{bin}}$ denote the voxel-level fine semantic, coarse-group, and binary occupancy losses, respectively. The term $\mathcal{L}^{p}_{\mathrm{plane}}$ denotes the projected binary occupancy loss on plane $p$. This auxiliary supervision directly constrains every plane to preserve the projected occupancy structure used to reconstruct the dense voxel grid.

\section{Simulation-to-Real Transfer}

Evaluation confined to simulation cannot establish whether knowledge learned from synthetic data remains effective in real-world scenes. To examine whether dense voxel supervision from SemanticVVSim benefits real-world semantic occupancy perception, we study simulation-to-real transfer. Conventional source-domain pretraining followed by target-domain fine-tuning does not explicitly separate transferable occupancy structure from source-specific observation patterns. Differences in scene appearance and sensor configurations between the simulation and real-world domains lead to a distribution shift in sensor observations. Therefore, direct observation-level alignment is unreliable.

Knowledge distillation transfers knowledge across domains without requiring aligned raw observations~\cite{gou2021kdsurvey,shu2021channelwise,park2019rkd}. However, direct distillation can propagate unreliable teacher outputs, while semantic-logit transfer requires compatible output spaces. Therefore, existing distillation approaches do not directly address spatially varying teacher reliability and incompatible semantic taxonomies in this setting.

To address these cross-domain transfer challenges, we introduce CHORD, which maps target-domain inputs to spatially aligned egocentric occupancy representations instead of aligning raw observations. A frozen \methodname{} teacher trained on SemanticVVSim and a target-domain student process each target sample. At the representation level, CHORD distills normalized plane features and spatial token relations, while target-domain supervision adapts the student to real observations. At the output level, confidence-guided distillation transfers only binary occupancy and compatible coarse semantic groups to accommodate differences between the source and target label spaces.

\subsection{Confidence-Aware Hierarchical Occupancy Distillation}

CHORD uses teacher binary confidence to select voxel locations for hierarchical logit distillation. $S$ denotes quantities from the student network and $T$ denotes quantities from the teacher network. Binary occupancy provides the selection signal because the free/occupied decision is shared across the source and target taxonomies and precedes coarse-group classification in the hierarchical decoder. Therefore, a confident binary decision identifies locations with a reliable occupancy state before the transfer of coarse semantic information. Let $\boldsymbol{\ell}^{\mathrm{bin},T}_{\nu}$ denote the binary logits at voxel $\nu$. We define
\begin{equation}
    M_{\nu}=\mathbb{1}\!\left[
    \max_{c\in\{\mathrm{free},\mathrm{occ}\}}
    \operatorname{softmax}(\boldsymbol{\ell}^{\mathrm{bin},T}_{\nu})_c
    \geq\tau_{\mathrm{conf}}\right],
    \label{eq:chord_confidence_mask}
\end{equation}
where $\tau_{\mathrm{conf}}$ is the confidence threshold. This mask retains locations where the teacher makes a confident binary decision and suppresses uncertain teacher outputs.

The binary confidence mask selects a common set of voxel locations for both binary-occupancy and coarse-group distillation. Let $q\in\{\mathrm{bin},\mathrm{grp}\}$ denote the binary-occupancy and coarse-group classification heads, respectively. At location $\nu$, $\boldsymbol{\ell}^{q,S}_{\nu}$ and $\boldsymbol{\ell}^{q,T}_{\nu}$ denote the logits for head $q$. The corresponding softened distributions are
\begin{equation}
    \begin{aligned}
        \boldsymbol{p}^{q,S}_{\nu}
        &=\operatorname{softmax}\!\left(
        \boldsymbol{\ell}^{q,S}_{\nu}/\tau_{\mathrm{KD}}\right),\\
        \boldsymbol{p}^{q,T}_{\nu}
        &=\operatorname{softmax}\!\left(
        \boldsymbol{\ell}^{q,T}_{\nu}/\tau_{\mathrm{KD}}\right).
    \end{aligned}
    \label{eq:chord_softened_distributions}
\end{equation}
where $\tau_{\mathrm{KD}}$ controls the smoothness of the distributions used for distillation. We normalize the loss by the number of selected locations:
\begin{equation}
    N_M=\max\!\left(1,\sum_{\nu\in\mathcal{G}}M_{\nu}\right).
    \label{eq:chord_mask_normalization}
\end{equation}
The masked distillation loss for head $q$ is
\begin{equation}
    \mathcal{L}_{q}^{\mathrm{KD}}
    =\frac{\tau_{\mathrm{KD}}^2}{N_M}
    \sum_{\nu\in\mathcal{G}}M_{\nu}\,
    \operatorname{KL}\!\left(
    \boldsymbol{p}^{q,T}_{\nu}\,\|\,
    \boldsymbol{p}^{q,S}_{\nu}\right).
    \label{eq:chord_masked_kl}
\end{equation}
Fine-grained semantic logits are excluded from distillation because the source and target taxonomies do not provide complete one-to-one correspondence between fine classes. Direct matching of these logits would align categories with different meanings and introduce invalid supervision. Instead, target-domain labels supervise the fine-grained student outputs.

\subsection{Latent Feature and Relational Distillation}

At the representation level, CHORD aligns the ABFI-updated plane features without requiring semantic labels or class-wise correspondence. Direct comparison can be dominated by differences in feature magnitude between the student and teacher networks. For each plane, CHORD removes this magnitude by normalizing every ABFI output token over the feature dimension:
\begin{equation}
    \begin{aligned}
        \mathbf{Z}^{p,S}_{i}
        &=\frac{\widetilde{\mathbf{C}}^{p,S}_{i}}
        {\max(\lVert\widetilde{\mathbf{C}}^{p,S}_{i}\rVert_2,\varepsilon)},\\
        \mathbf{Z}^{p,T}_{i}
        &=\frac{\widetilde{\mathbf{C}}^{p,T}_{i}}
        {\max(\lVert\widetilde{\mathbf{C}}^{p,T}_{i}\rVert_2,\varepsilon)},
    \end{aligned}
    \label{eq:chord_token_normalization}
\end{equation}
where $i$ indexes plane tokens. The constant $\varepsilon$ provides numerical stability. This operation retains the relative activation pattern across feature channels. The latent loss is
\begin{equation}
    \mathcal{L}_{\mathrm{lat}}
    =\frac{1}{3}\sum_{p\in\mathcal{P}}
    \operatorname{SmoothL1}(\mathbf{Z}^{p,S},\mathbf{Z}^{p,T}).
    \label{eq:chord_latent_loss}
\end{equation}
We use Smooth L1 loss to align the normalized student and teacher features. Its linear growth for large residuals prevents substantial cross-domain discrepancies from dominating feature alignment.

CHORD further aligns relations between spatial tokens. Voxel states exhibit spatial dependencies that encode scene structure. These dependencies are reflected in relations between plane tokens. Pointwise feature alignment supervises each token independently but does not explicitly constrain the pairwise geometry of the token set. Therefore, we construct a relation matrix on each plane:
\begin{equation}
    \mathbf{R}^{p,S}=\mathbf{Z}^{p,S}(\mathbf{Z}^{p,S})^{\top},\qquad
    \mathbf{R}^{p,T}=\mathbf{Z}^{p,T}(\mathbf{Z}^{p,T})^{\top}.
    \label{eq:chord_relation_matrix}
\end{equation}
Each $\mathbf{R}^{p}$ is the Gram matrix of the normalized tokens on plane $p$. The entry $R^{p}_{ij}=(\mathbf{Z}^{p}_{i})^{\top}\mathbf{Z}^{p}_{j}$ equals the cosine similarity between tokens $i$ and $j$. For tokens with pre-normalization norms no smaller than $\varepsilon$, Eq.~\eqref{eq:chord_token_normalization} gives $\lVert\mathbf{Z}^{p}_{i}\rVert_2=\lVert\mathbf{Z}^{p}_{j}\rVert_2=1$. The squared Euclidean distance between two normalized tokens can then be expanded as
\begin{equation}
    \begin{aligned}
        \lVert\mathbf{Z}^{p}_{i}-\mathbf{Z}^{p}_{j}\rVert_2^2
        &=\lVert\mathbf{Z}^{p}_{i}\rVert_2^2
        +\lVert\mathbf{Z}^{p}_{j}\rVert_2^2
        -2(\mathbf{Z}^{p}_{i})^{\top}\mathbf{Z}^{p}_{j} \\
        &=2-2R^{p}_{ij}.
    \end{aligned}
    \label{eq:chord_relation_distance}
\end{equation}
Therefore, the Gram matrix encodes pairwise distances among normalized tokens on each plane. Each matrix entry links two spatial locations, so the relation loss supervises the similarity pattern across the complete sampled token set. The latent loss aligns corresponding token features, while the relation loss constrains how each token relates to other sampled tokens. Together, the two losses supervise both token-level features and relations across spatial locations. CHORD constructs the relation matrices from uniformly sampled plane tokens and minimizes
\begin{equation}
    \mathcal{L}_{\mathrm{rel}}
    =\frac{1}{3}\sum_{p\in\mathcal{P}}
    \operatorname{SmoothL1}(\mathbf{R}^{p,S},\mathbf{R}^{p,T}).
    \label{eq:chord_relation_loss}
\end{equation}
The complete CHORD training objective is
\begin{equation}
    \begin{aligned}
    \mathcal{L}_{\mathrm{CHORD}}
    =\mathcal{L}_{\mathrm{sup}}
    &+\lambda_{\mathrm{lat}}\mathcal{L}_{\mathrm{lat}}
    +\lambda_{\mathrm{rel}}\mathcal{L}_{\mathrm{rel}}\\
    &+\lambda_{\mathrm{bin}}\mathcal{L}^{\mathrm{KD}}_{\mathrm{bin}}
    +\lambda_{\mathrm{grp}}\mathcal{L}^{\mathrm{KD}}_{\mathrm{grp}}.
    \end{aligned}
    \label{eq:chord_total_loss}
\end{equation}
The teacher remains frozen. CHORD transfers latent features, token relations, binary occupancy, and coarse groups without distilling incompatible fine classes.

\section{Experiments}
\label{sec:experiments}

We evaluate \methodname{} under both single-agent and multi-agent settings on SemanticVVSim. We report quantitative results and component ablations on this benchmark and further assess simulation-to-real transfer on Occ3D-nuScenes~\cite{caesar2020nuscenes,tian2023occ3d}.

\begin{table*}[!t]
\centering
\caption{Single-agent semantic occupancy results (\%) on SemanticVVSim.}
\label{tab:single_agent_miou}
\setlength{\tabcolsep}{2.2pt}
\renewcommand{\arraystretch}{1.05}
\resizebox{\textwidth}{!}{%
\begin{tabular}{lcccccccccccccccc}
\hline
Method & Geo. IoU & mIoU & \freeclassbox Free & \classbox{svBarrier} Barr. & \classbox{svCar} Car & \classbox{svPedestrian} Ped. & \classbox{svTruck} Truck & \classbox{svSpeedLimitSign} S.L. Sign & \classbox{svVan} Van & \classbox{svTrailer} Trailer & \classbox{svTrafficCone} T. Cone & \classbox{svDrivable} Driv. Surf. & \classbox{svTerrain} Terrain & \classbox{svSidewalk} Sidew. & \classbox{svManmade} Man. & \classbox{svVegetation} Veg. \\
\hline
PointPillars$^\dagger$~\cite{lang2019pointpillars} & 37.60 & 10.53 & 80.62 & 0.22 & 0.11 & 0.00 & 0.02 & 0.00 & 0.00 & 0.00 & 0.10 & 38.62 & 0.11 & 19.42 & 6.97 & 1.24 \\
SurroundOcc~\cite{wei2023surroundocc} & 49.62 & 20.94 & 83.26 & 9.68 & 13.16 & 0.28 & 1.77 & 0.01 & 0.59 & 2.47 & 0.51 & 55.45 & 37.13 & 42.45 & 29.42 & 16.91 \\
MonoScene~\cite{cao2022monoscene} & 59.27 & 29.83 & 88.33 & 33.84 & 41.71 & 5.89 & 1.46 & 0.00 & 0.14 & 0.00 & 0.17 & 80.35 & 48.31 & 42.80 & 35.90 & 38.73 \\
CoHFF~\cite{song2024cohff} & 67.63 & 42.52 & 89.62 & 49.72 & 48.79 & 7.60 & 9.15 & 4.18 & 23.28 & 4.48 & 27.06 & 90.23 & 68.08 & 70.77 & 54.31 & 47.94 \\
VoxelNet$^\dagger$~\cite{zhou2018voxelnet} & 61.87 & 44.51 & 84.53 & 54.73 & 55.77 & \underline{35.45} & \underline{36.73} & 24.26 & 22.40 & \textbf{21.86} & 60.54 & 62.15 & 46.46 & 27.83 & 38.26 & 52.13 \\
BEVFusion$^\dagger$~\cite{liu2023bevfusion} & 64.30 & 46.28 & 90.39 & \underline{58.65} & 64.32 & 29.63 & 32.76 & \textbf{26.19} & \underline{24.30} & 18.09 & 55.64 & 65.65 & 52.34 & 36.83 & 37.22 & 55.86 \\
VOGS-CP~\cite{chen2026vogscp} & 70.83 & 47.84 & 91.64 & 46.96 & 56.32 & 5.26 & 21.78 & 2.74 & 20.27 & 9.24 & 64.58 & 92.08 & 75.06 & 74.69 & 53.40 & 55.75 \\
OccCylindrical~\cite{ming2025occcylindrical} & 69.55 & 49.25 & 91.33 & \textbf{65.30} & 59.41 & 33.51 & 26.44 & \underline{24.55} & 12.02 & 13.61 & 28.17 & 84.97 & 69.89 & 68.26 & 50.77 & \underline{61.28} \\
TPVFormer~\cite{huang2023tpvformer} & \underline{73.69} & \underline{53.54} & \underline{92.50} & 50.77 & \underline{65.09} & 16.29 & 30.16 & 2.70 & 23.44 & 5.13 & \textbf{87.75} & \underline{95.06} & \underline{78.53} & \underline{82.30} & \underline{59.92} & 59.87 \\
\textbf{\methodname{}} & \textbf{75.47} & \textbf{59.55} & \textbf{93.06} & 55.40 & \textbf{71.61} & \textbf{39.44} & \textbf{37.75} & 18.89 & \textbf{29.80} & \underline{20.57} & \underline{81.64} & \textbf{95.34} & \textbf{79.87} & \textbf{84.03} & \textbf{61.29} & \textbf{64.97} \\
\hline
\end{tabular}%
}
\begin{minipage}{\textwidth}
\footnotesize
Bold and underlined values indicate the best and second-best results, respectively.
$^\dagger$ denotes a benchmark implementation that uses the corresponding feature encoder with a semantic occupancy head.
\end{minipage}
\end{table*}

\subsection{Experimental Setup}

\subsubsection{Datasets and Task Settings}

SemanticVVSim contains 61,625 frames. Each frame represents one synchronized timestamp and contains observations from all available agents together with the associated semantic occupancy ground truth. We divide the dataset into 38,000 training frames, 5,125 validation frames, and 18,500 test frames. The split maintains a balanced distribution of traffic scenarios and environmental conditions across the three subsets. The occupancy volume spans $[-54,54]$~m along both horizontal axes and $[-2,4]$~m along the vertical axis. The voxel size is $0.6$~m. The multi-agent setting uses a communication range of $70$~m.

\subsubsection{Evaluation Metrics}

We use class-wise intersection over union (IoU) and mean IoU (mIoU) for semantic occupancy evaluation. SemanticVVSim defines 14 evaluation classes: free, barrier, car, pedestrian, truck, speed limit sign, van, trailer, traffic cone, drivable surface, terrain, sidewalk, manmade, and vegetation. All voxels with valid evaluation labels within the occupancy volume participate in evaluation. No visibility mask is applied. For evaluation class $c$, IoU is defined as
\begin{equation}
    \operatorname{IoU}_{c}
    =\frac{\operatorname{TP}_{c}}
    {\operatorname{TP}_{c}+\operatorname{FP}_{c}+\operatorname{FN}_{c}}.
\end{equation}
Here, $\operatorname{TP}_{c}$ denotes the number of voxels correctly classified as class $c$. $\operatorname{FP}_{c}$ counts voxels from other classes classified as class $c$. $\operatorname{FN}_{c}$ counts voxels of class $c$ classified as another class.
We compute mIoU as the arithmetic mean of the IoUs over all 14 evaluation classes.

For geometric occupancy evaluation, we merge the 13 non-free classes into one occupied class and treat free as the negative class. Let $\operatorname{TP}_{\mathrm{occ}}$, $\operatorname{FP}_{\mathrm{occ}}$, and $\operatorname{FN}_{\mathrm{occ}}$ denote the binary counts for occupied. We report the IoU of occupied voxels as geometric IoU. The metric is computed as $\operatorname{IoU}_{\mathrm{geo}}=\operatorname{TP}_{\mathrm{occ}}/(\operatorname{TP}_{\mathrm{occ}}+\operatorname{FP}_{\mathrm{occ}}+\operatorname{FN}_{\mathrm{occ}})$.

\subsubsection{Compared Methods}

We compare \methodname{} with nine representative baselines that cover LiDAR-only, camera-only, and LiDAR-camera settings. The LiDAR-only group includes \textbf{PointPillars}~\cite{lang2019pointpillars} and \textbf{VoxelNet}~\cite{zhou2018voxelnet}, which use pillar-based and voxel-based encoders, respectively. The camera-only group includes \textbf{MonoScene}~\cite{cao2022monoscene}, \textbf{SurroundOcc}~\cite{wei2023surroundocc}, \textbf{TPVFormer}~\cite{huang2023tpvformer}, \textbf{CoHFF}~\cite{song2024cohff}, and \textbf{VOGS-CP}~\cite{chen2026vogscp}. These methods use a dense volume, multi-scale volumes, tri-perspective planes, hybrid task and orthogonal attention features, and sparse 3D semantic Gaussians, respectively. The LiDAR-camera group includes \textbf{BEVFusion}~\cite{liu2023bevfusion} and \textbf{OccCylindrical}~\cite{ming2025occcylindrical}, which fuse sensor features in Cartesian BEV and cylindrical tri-perspective representations. For multi-agent evaluation, CoHFF and VOGS-CP use the respective native cross-agent fusion designs~\cite{song2024cohff,chen2026vogscp}. Each of the remaining seven baselines uses a representation-compatible EQCA module to aggregate spatially aligned features from all valid agents.

\subsubsection{Implementation Details}

\methodname{} and the seven baselines equipped with EQCA use the same training settings. CoHFF and VOGS-CP retain their original cross-agent fusion modules. Training uses AdamW with a base learning rate of $1\times 10^{-4}$ and weight decay of $0.01$. A linear warm-up increases the learning rate from $10^{-3}$ of the base value during the first 1,000 iterations. A multi-step schedule reduces the learning rate by a factor of $0.1$ at epochs 20 and 23. Training runs for 24 epochs on four NVIDIA GeForce RTX 4090 GPUs with one sample per GPU, resulting in a total batch size of four. Gradient clipping limits the $\ell_2$ norm to 35.

\subsection{Quantitative Analysis of Single-Agent Performance}

We first compare the evaluated methods in the single-agent setting. Table~\ref{tab:single_agent_miou} reports the class-wise IoU and overall results. \methodname{} achieves the best overall performance among these methods, with 59.55\% mIoU and 75.47\% geometric IoU. More importantly, the class-wise results show that the advantage covers both foreground traffic participants and static scene regions instead of concentrating on a small subset of classes.

For foreground traffic participants, \methodname{} obtains the highest IoU for car, pedestrian, truck, and van. These categories cover road users with different spatial scales and shapes. The consistent results indicate strong semantic occupancy perception for safety-relevant foreground objects. \methodname{} also ranks first for drivable surface, terrain, sidewalk, manmade, and vegetation. These classes describe large static surfaces and scene structures. The broad class-wise advantage shows that \methodname{} recognizes foreground traffic participants while retaining accurate estimates of the surrounding scene layout.

\subsection{Quantitative Analysis of Multi-Agent Performance}

We then compare \methodname{} with the other nine methods in the multi-agent setting. Table~\ref{tab:multi_agent_miou} reports the class-wise IoU and overall results. All ten methods achieve higher overall mIoU through cooperation. This consistent improvement shows that complementary observations benefit semantic occupancy perception.

\begin{table*}[!t]
\centering
\caption{Multi-agent semantic occupancy results (\%) on SemanticVVSim.}
\label{tab:multi_agent_miou}
\setlength{\tabcolsep}{2.2pt}
\renewcommand{\arraystretch}{1.05}
\resizebox{\textwidth}{!}{%
\begin{tabular}{lcccccccccccccccc}
\hline
Method & Geo. IoU & mIoU & \freeclassbox Free & \classbox{svBarrier} Barr. & \classbox{svCar} Car & \classbox{svPedestrian} Ped. & \classbox{svTruck} Truck & \classbox{svSpeedLimitSign} S.L. Sign & \classbox{svVan} Van & \classbox{svTrailer} Trailer & \classbox{svTrafficCone} T. Cone & \classbox{svDrivable} Driv. Surf. & \classbox{svTerrain} Terrain & \classbox{svSidewalk} Sidew. & \classbox{svManmade} Man. & \classbox{svVegetation} Veg. \\
\hline
PointPillars$^\dagger$~\cite{lang2019pointpillars} & 43.76 & 15.30 & 82.45 & 9.51 & 1.24 & 0.02 & 0.33 & 0.01 & 0.02 & 0.03 & 0.77 & 46.50 & 23.31 & 27.50 & 19.89 & 2.60 \\
SurroundOcc~\cite{wei2023surroundocc} & 52.53 & 23.35 & 85.02 & 17.08 & 15.49 & 0.25 & 1.59 & 0.01 & 0.87 & 2.24 & 1.54 & 62.94 & 41.79 & 46.44 & 32.93 & 18.70 \\
MonoScene~\cite{cao2022monoscene} & 65.27 & 34.01 & 90.03 & 38.03 & 44.23 & 4.72 & 2.42 & 0.00 & 0.28 & 0.00 & 0.02 & 84.08 & 66.00 & 55.81 & 45.61 & 44.90 \\
VoxelNet$^\dagger$~\cite{zhou2018voxelnet} & 62.10 & 48.02 & 87.74 & 58.91 & 58.41 & \underline{39.32} & \underline{41.22} & \underline{27.21} & 26.00 & \textbf{25.00} & 63.10 & 65.87 & 49.94 & 32.68 & 40.43 & 56.46 \\
CoHFF~\cite{song2024cohff} & 69.77 & 48.79 & 89.20 & 54.09 & 52.05 & 21.44 & 21.32 & 20.80 & \underline{31.73} & 19.81 & 33.64 & 89.52 & 70.01 & 72.03 & 57.19 & 50.23 \\
BEVFusion$^\dagger$~\cite{liu2023bevfusion} & 70.45 & 49.85 & 93.11 & \underline{62.01} & 68.89 & 33.47 & 36.95 & \textbf{29.61} & 29.21 & 21.07 & 59.30 & 68.92 & 56.79 & 40.76 & 39.76 & 58.07 \\
VOGS-CP~\cite{chen2026vogscp} & 73.04 & 51.21 & 92.22 & 48.44 & 59.46 & 12.17 & 30.14 & 4.48 & 24.99 & 2.04 & 77.06 & 93.28 & 75.71 & 80.25 & 58.99 & 57.72 \\
OccCylindrical~\cite{ming2025occcylindrical} & 72.31 & 52.67 & \textbf{94.19} & \textbf{69.72} & 62.56 & 37.49 & 30.08 & 27.02 & 16.53 & 16.97 & 32.36 & 86.75 & 72.61 & 70.51 & 54.36 & \underline{66.28} \\
TPVFormer~\cite{huang2023tpvformer} & \underline{74.67} & \underline{56.33} & 92.85 & 51.18 & \underline{74.41} & 25.01 & 38.14 & 1.47 & 25.78 & 6.83 & \textbf{90.70} & \underline{96.16} & \underline{79.14} & \underline{84.52} & \underline{61.28} & 61.13 \\
\textbf{\methodname{}} & \textbf{76.57} & \textbf{62.75} & \underline{93.43} & 55.75 & \textbf{79.31} & \textbf{52.91} & \textbf{46.21} & 21.48 & \textbf{34.18} & \underline{23.78} & \underline{79.54} & \textbf{96.35} & \textbf{79.47} & \textbf{85.30} & \textbf{62.29} & \textbf{68.46} \\
\hline
\end{tabular}%
}
\begin{minipage}{\textwidth}
\footnotesize
Bold and underlined values indicate the best and second-best results, respectively.
$^\dagger$ denotes a benchmark implementation that uses the corresponding feature encoder with a semantic occupancy head.
\end{minipage}
\end{table*}

\methodname{} achieves the best overall multi-agent performance among the evaluated methods, with 62.75\% mIoU and 76.57\% geometric IoU. The class-wise results show that the benefits of multi-agent cooperation vary across categories rather than improving all categories uniformly. Compared with the single-agent setting, the main gains concentrate on foreground traffic participants, including cars, pedestrians, trucks, and vans. This pattern is consistent with the role of multi-view observations. Foreground objects are often affected by occlusion and limited visibility, while observations from other agents can supplement target regions missing from a single view. By contrast, large static regions, such as drivable surfaces, terrain, and sidewalks, are typically observed more completely in the single-agent setting, leaving less room for improvement through cooperation. Nevertheless, \methodname{} remains the top-performing method for major static-scene categories, showing that the foreground gains coexist with leading performance in overall scene layout modeling.

More importantly, the benefit of EQCA is not limited to \methodname{}. All seven baselines equipped with representation-compatible EQCA modules outperform the corresponding single-agent models. These baselines cover LiDAR-only, camera-only, and LiDAR-camera inputs, together with pillar, voxel, BEV, dense-volume, and tri-perspective representations. The consistent gains across different model families show that EQCA supports multi-agent feature aggregation across diverse encoders and spatial representations.

\subsection{Robustness Analysis}

We evaluate \methodname{} under two practical perturbations in cooperative perception: communication delay and pose error. The zero-perturbation result provides a common reference.

\begin{table}[t]
\centering
\caption{\methodname{} results (\%) under different communication delays on SemanticVVSim.}
\label{tab:delay_robustness}
\setlength{\tabcolsep}{9pt}
\renewcommand{\arraystretch}{1.05}
\begin{tabular}{ccc}
\hline
Delay (ms) & Geo. IoU & mIoU \\
\hline
0 & 76.57 & 62.75 \\
100 & 76.55 & 62.52 \\
200 & 76.55 & 62.19 \\
300 & 76.53 & 61.65 \\
400 & 76.52 & 61.34 \\
500 & 76.51 & 61.04 \\
600 & 76.50 & 60.90 \\
700 & 76.48 & 60.67 \\
800 & 76.47 & 60.53 \\
900 & 76.46 & 60.37 \\
1000 & 76.46 & 60.29 \\
\hline
\end{tabular}
\end{table}

\begin{table}[t]
\centering
\caption{\methodname{} results (\%) under Gaussian pose perturbations on SemanticVVSim.}
\label{tab:pose_robustness}
\setlength{\tabcolsep}{5pt}
\renewcommand{\arraystretch}{1.05}
\begin{tabular}{cccc}
\hline
$\sigma_{xyz}$ (m) & $\sigma_{\mathrm{yaw}}$ ($^\circ$) & Geo. IoU & mIoU \\
\hline
0.0 & 0.0 & 76.57 & 62.75 \\
0.2 & 0.4 & 76.31 & 61.27 \\
0.5 & 1.0 & 75.79 & 59.72 \\
0.5 & 2.0 & 75.58 & 59.47 \\
1.0 & 5.0 & 75.13 & 58.48 \\
\hline
\end{tabular}
\end{table}

Communication delay leads to a gradual change in performance. As the delay increases from 0 to 1000~ms, geometric IoU remains between 76.46\% and 76.57\%. Over the same interval, mIoU decreases from 62.75\% to 60.29\%. The stable geometric IoU shows that \methodname{} maintains consistent occupied-space estimates under delayed collaborative observations. The limited decrease in mIoU further indicates that semantic occupancy perception remains stable as the delay increases.

\methodname{} also shows smooth performance changes under pose perturbations. As the combined translation and yaw perturbations increase, geometric IoU changes from 76.57\% to 75.13\%, and mIoU changes from 62.75\% to 58.48\%. No perturbation level causes an abrupt performance drop. Under the largest pose perturbation, \methodname{} still retains 75.13\% geometric IoU and 58.48\% mIoU, showing tolerance to spatial misalignment.

Overall, \methodname{} preserves most of the zero-perturbation performance under both communication delay and pose error. The gradual performance changes across both tests show that the cooperative fusion framework maintains stable semantic occupancy estimates under imperfect temporal and spatial alignment.

\subsection{Ablation Studies}

We compare the complete \methodname{} model with four variants under both single-agent and multi-agent settings. The variants examine the two modality branches, ABFI, and hierarchical occupancy decoding. The single-modality variants keep only the camera or LiDAR branch. ABFI is disabled in these variants because ABFI requires both modalities. The ABFI ablation keeps both branches but uses simple feature fusion. Simple fusion projects sparse LiDAR voxel features onto three orthogonal camera planes and averages voxels at the same location. A plane-specific linear layer fuses the features, followed by residual addition and LayerNorm. The decoder ablation also keeps both branches and replaces the hierarchical decoder with a flat classifier.

\begin{table}[t]
\centering
\caption{Component ablations of \methodname{} on SemanticVVSim under single-agent and multi-agent settings.}
\label{tab:comaf_occ_ablation}
\setlength{\tabcolsep}{9pt}
\renewcommand{\arraystretch}{1.05}
\begin{tabular}{lcc}
\hline
Model Variant & Geo. IoU & mIoU \\
\hline
\multicolumn{3}{l}{\textit{(a) Single-Agent}} \\
Full model & \textbf{75.47} & \textbf{59.55} \\
w/o camera branch & 71.20 & 47.24 \\
w/o LiDAR branch & 74.65 & 56.92 \\
w/o ABFI & 74.90 & 57.50 \\
w/o hierarchical decoder & 75.28 & 57.63 \\
\hline
\multicolumn{3}{l}{\textit{(b) Multi-Agent}} \\
Full model & \textbf{76.57} & \textbf{62.75} \\
w/o camera branch & 72.49 & 50.70 \\
w/o LiDAR branch & 74.89 & 59.09 \\
w/o ABFI & 75.39 & 60.04 \\
w/o hierarchical decoder & 74.64 & 59.46 \\
\hline
\end{tabular}
\end{table}

Using either modality alone reduces performance under both the single-agent and multi-agent settings, while the complete model consistently achieves the best results. This pattern demonstrates the complementary value of camera and LiDAR features. The larger performance drop caused by removing the camera branch indicates that camera features provide rich semantic information. LiDAR features provide complementary geometric and spatial information. Therefore, the fusion of the two modalities improves overall performance.

In both settings, removing ABFI affects semantic mIoU more strongly than geometric IoU. This result indicates that bidirectional interaction mainly enhances semantic feature fusion rather than only distinguishing free space from occupied space. Removing the hierarchical decoder shows a similar trend, with a larger effect on semantic classification. Therefore, the hierarchical decoder helps convert fused representations into fine-grained semantic estimates.

\subsection{Efficiency Analysis}

\begin{table*}[t]
\centering
\caption{Model complexity and inference throughput under single-agent and multi-agent settings. Native FPS denotes throughput without TensorRT acceleration.}
\label{tab:efficiency_analysis}
\setlength{\tabcolsep}{9pt}
\renewcommand{\arraystretch}{1.05}
\begin{tabular}{lcccccc}
\hline
Method & \multicolumn{2}{c}{Parameters (M) $\downarrow$} & \multicolumn{2}{c}{Native FPS $\uparrow$} & \multicolumn{2}{c}{TensorRT FPS $\uparrow$} \\
\cline{2-3}\cline{4-5}\cline{6-7}
& Single & Multi & Single & Multi & Single & Multi \\
\hline
PointPillars$^\dagger$~\cite{lang2019pointpillars} & 4.22 & 5.41 & 11.85 & 9.83 & 41.52 & 28.44 \\
SurroundOcc~\cite{wei2023surroundocc} & 189.70 & 192.47 & 5.76 & 4.93 & 10.25 & 8.60 \\
MonoScene~\cite{cao2022monoscene} & 65.19 & 65.33 & 4.68 & 3.25 & 12.20 & 7.37 \\
CoHFF~\cite{song2024cohff} & 51.41 & 57.43 & 10.97 & 7.51 & 18.50 & 12.38 \\
VoxelNet$^\dagger$~\cite{zhou2018voxelnet} & 10.20 & 11.38 & 10.00 & 5.86 & 27.06 & 9.95 \\
BEVFusion$^\dagger$~\cite{liu2023bevfusion} & 42.55 & 43.73 & 8.08 & 4.07 & 19.82 & 7.20 \\
VOGS-CP~\cite{chen2026vogscp} & 75.61 & 75.67 & 9.65 & 3.07 & 17.69 & 10.55 \\
OccCylindrical~\cite{ming2025occcylindrical} & 82.28 & 82.68 & 6.70 & 2.95 & 9.44 & 4.92 \\
TPVFormer~\cite{huang2023tpvformer} & 77.31 & 78.89 & 6.66 & 5.47 & 18.82 & 13.59 \\
\textbf{\methodname{}} & 76.25 & 77.14 & 5.53 & 4.51 & 18.26 & 13.73 \\
\hline
\end{tabular}
\end{table*}

\begin{table*}[!t]
\centering
\caption{Class-wise \methodname{} simulation-to-real transfer results (\%) on Occ3D-nuScenes. Source pretraining uses SemanticVVSim. Bold values indicate the best result in each column.}
\label{tab:sim2real}
\setlength{\tabcolsep}{2pt}
\renewcommand{\arraystretch}{1.05}
\resizebox{\textwidth}{!}{%
\begin{tabular}{lccccccccccccccccccc}
\hline
Training setting & Geo. IoU & mIoU & Free & Barr. & Bicy. & Bus & Car & Constr. Veh. & Motor. & Ped. & Truck & Trailer & T. Cone & Driv. Surf. & Other Flat & Terrain & Sidew. & Man. & Veg. \\
\hline
Direct target training & 81.68 & 46.05 & 92.60 & 51.02 & 19.43 & 44.21 & 49.04 & 23.67 & 25.26 & 41.87 & 37.71 & 40.37 & 25.22 & \textbf{79.13} & 37.34 & 50.31 & 47.24 & 59.15 & 59.33 \\
Pretraining + target fine-tuning & 81.64 & 46.41 & \textbf{92.67} & 52.08 & 19.13 & 46.37 & 50.23 & 25.67 & 23.17 & \textbf{42.60} & 38.57 & 38.91 & \textbf{26.64} & 78.06 & 38.14 & 50.49 & 47.01 & \textbf{59.21} & \textbf{60.04} \\
Pretraining + CHORD fine-tuning & \textbf{81.71} & \textbf{48.40} & 92.25 & \textbf{53.13} & \textbf{25.57} & \textbf{46.89} & \textbf{53.99} & \textbf{28.33} & \textbf{36.89} & 42.29 & \textbf{43.33} & \textbf{41.72} & 25.22 & 79.12 & \textbf{38.64} & \textbf{51.38} & \textbf{48.33} & 57.39 & 58.33 \\
\hline
\end{tabular}%
}
\end{table*}

Table~\ref{tab:efficiency_analysis} compares model complexity and inference throughput under the single-agent and multi-agent settings, with a focus on the balance between perception accuracy and deployment efficiency. In the multi-agent setting, to simulate real-world cooperative computation across agents, feature extraction for different agents runs in parallel on separate GPUs, and the intermediate features are aggregated during cooperative fusion. \methodname{} uses a moderate parameter budget compared with other dense occupancy models. More importantly, the small parameter gap between the single-agent and multi-agent settings shows that cooperative perception introduces only limited additional model complexity. The accuracy gains from multi-agent observations therefore do not depend on a large expansion of model capacity.

Without TensorRT acceleration, \methodname{} maintains usable inference throughput under both the single-agent and multi-agent settings. Although several lightweight baselines run faster, the corresponding occupancy accuracy remains substantially lower. \methodname{} therefore provides a stronger balance between perception quality and inference efficiency.

\begin{figure}[!t]
\centering
\includegraphics[width=\columnwidth]{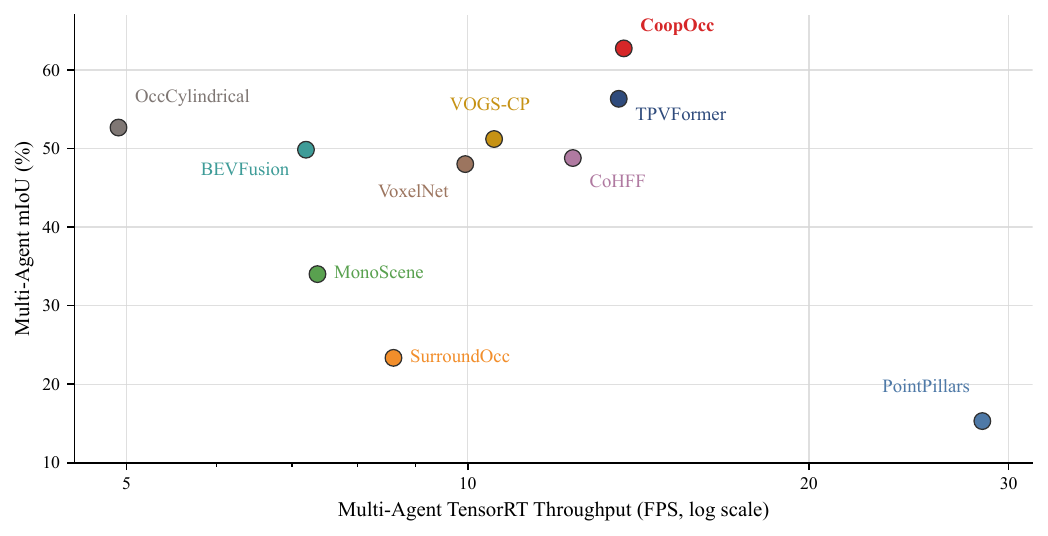}
\caption{Multi-agent accuracy-throughput trade-off with TensorRT acceleration.}
\label{fig:efficiency_tradeoff}
\end{figure}

TensorRT further improves the inference throughput of \methodname{} under both settings. The multi-agent accuracy-throughput trade-off under TensorRT acceleration is shown in Figure~\ref{fig:efficiency_tradeoff}. This result shows that the model benefits from deployment-oriented optimization and has practical potential for online cooperative perception.

\subsection{Simulation-to-Real Transfer}

Table~\ref{tab:sim2real} compares three training settings of the same \methodname{} architecture on Occ3D-nuScenes~\cite{caesar2020nuscenes,tian2023occ3d}.
These settings comprise direct target-domain training, SemanticVVSim pretraining followed by target-domain fine-tuning, and SemanticVVSim pretraining followed by target-domain fine-tuning with CHORD.

The three training settings differ mainly in semantic recognition rather than geometric occupancy estimation. Geometric IoU remains nearly unchanged across the three settings, while the semantic results differ more clearly. This separation indicates that simulation-to-real transfer mainly affects class discrimination within the occupied space rather than free-space and occupied-space separation.

Ordinary SemanticVVSim pretraining yields essentially the same overall performance as direct target training. The class-wise changes also lack a consistent direction. Therefore, source initialization alone does not provide a meaningful transfer benefit under changes in sensor observations and label spaces.

By contrast, CHORD produces the strongest mIoU while preserving geometric IoU. The largest improvements over direct target training occur for motorcycle, bicycle, truck, car, and construction vehicle. Together with the stable geometric IoU, these class-wise improvements indicate that CHORD preserves occupancy geometry while improving fine-grained semantic discrimination within occupied regions.

This behavior is consistent with the CHORD design. Confidence masking limits supervision from uncertain teacher outputs. Hierarchical and relational distillation transfers shared occupancy structure without directly matching incompatible fine-grained categories. Overall, CHORD improves fine-grained semantic recognition while preserving geometric occupancy under simulation-to-real transfer.

\section{Conclusion}
\label{sec:conclusion}

This paper studies cooperative 3D semantic occupancy perception through multi-agent and cross-modal fusion. To support research in this direction, we construct SemanticVVSim, a multi-agent, multi-modal dataset for semantic occupancy perception, and establish a unified benchmark. \methodname{} combines camera semantics and LiDAR geometry through anchor-based bidirectional feature interaction, while ego-query cross-agent attention aggregates spatially aligned information from collaborating agents. To address domain and label-space differences between simulation and real-world data, CHORD transfers knowledge through confidence-aware hierarchical and relational distillation. Experimental results show that multi-agent cooperation provides positive overall gains for all ten evaluated methods. The results also demonstrate effective multi-agent and cross-modal fusion in \methodname{} and improved semantic occupancy performance on Occ3D-nuScenes through CHORD.

\bibliographystyle{IEEEtran}
\bibliography{ref}

\vfill

\end{document}